\documentclass[11pt, a4paper, copyright, goog]{google}

\usepackage[authoryear, sort&compress, round]{natbib}
\usepackage{graphicx}
\usepackage{amsmath}
\usepackage{booktabs} % For professional tables
\usepackage{caption}
\usepackage{subcaption} % For subfigures
\usepackage{textcomp} % For text symbols like \textmu
\usepackage{gensymb} % For °
\usepackage{array}   % For table column formatting
\usepackage{fancyhdr} % For custom headers/footers
\usepackage{authblk} % The recommended package for author affiliations
\usepackage{booktabs} % For \toprule, \midrule, \bottomrule
\usepackage{tabularx}   % For the tabularx environment
\usepackage{gensymb}    % For the ° command
\usepackage{enumitem}
\usepackage{xurl} 
\usepackage{ragged2e}
\uselogo{} 

\title{An Operational Deep Learning System for Satellite-Based High-Resolution Global Nowcasting}

\correspondingauthor{shreyaa@google.com}

\renewcommand{\today}{2025-10-10}

\author[1]{Shreya Agrawal}
\author[1]{Mohammed Alewi Hassen}
\author[1]{Emmanuel Asiedu Brempong}
\author[1]{Boris Babenko}
\author[1]{Fred Zyda}
\author[1]{Olivia Graham}
\author[2]{Di Li}
\author[1]{Samier Merchant}
\author[1]{Santiago Hincapie Potes}
\author[1]{Tyler Russell}
\author[3]{Danny Cheresnick}
\author[3]{Aditya Prakash Kakkirala}
\author[2]{Stephan Rasp}
\author[1]{Avinatan Hassidim}
\author[1]{Yossi Matias}
\author[1]{Nal Kalchbrenner}
\author[1]{Pramod Gupta}
\author[1]{Jason Hickey}
\author[1]{Aaron Bell}
\affil[1]{\thepa{} Research}
\affil[2]{\thepa{} DeepMind}
\affil[3]{\thepa{} Search}

\begin{abstract}
Precipitation nowcasting, which predicts rainfall up to a few hours ahead, is a critical tool for vulnerable communities in the Global South that are frequently exposed to intense, rapidly developing storms. For these regions, timely forecasts provide a crucial window to protect lives and livelihoods. Traditional numerical weather prediction (NWP) methods often suffer from high latencies, low spatial and temporal resolutions, and significant gaps in accuracy across the world.  Recent progress in machine learning-based nowcasting methods, commonly used in the Global North, cannot be extended to the Global South due to extremely sparse radar coverage. Here, we present Global MetNet, an operationally ready global machine learning nowcasting model. It primarily leverages the Global Precipitation Mission’s (GPM) CORRA dataset and geostationary satellite data, along with global NWP data, to predict precipitation for the next 12 hours. The model operates at a high resolution of approximately 0.05$^{\circ} (\sim$5km) spatially and 15 minutes temporally. Global MetNet significantly outperforms industry-standard hourly forecasts and achieves a significantly higher skill, making the forecasts useful in a much larger area of the world than previously available. Our model demonstrates better skill in data-sparse regions than even the best high-resolution NWP models achieve in the US. Validated using ground radar and satellite data, it shows significant improvements across key metrics like the critical success index and fractions skill score for all precipitation rates and lead times. Crucially, our model operates under real-time conditions and generates forecasts in under a minute, making it readily deployable for diverse applications. It is already deployed for millions of users on Google Search. This work represents a key step in reducing global disparities in forecast quality and integrating sparse, high-resolution satellite observations into weather forecasting.
\end{abstract}

\begin{document}

\maketitle

\section{Introduction}
\label{sec:introduction}

% Precipitation nowcasting focuses on predicting precipitation events in the very near future, typically from the present moment up to a few hours ahead.  While it provides crucial information for numerous applications, its true urgency is demonstrated in regions in the Global South frequently exposed to intense, rapidly developing storms. For vulnerable communities in the tropics, timely nowcasts are often the last line of defense. They provide the crucial window needed to protect smallholder farms from torrential downpours, secure informal market stalls from flash floods, and enable life-saving evacuations. However, the chaotic nature of rapidly intensifying convective storms and high precipitation rates in these regions make producing accurate nowcasts a significant scientific challenge, one that must be met to mitigate disproportionate impacts on life and livelihood.
Nowcasting, the ability to forecast detailed local weather conditions from the present up to a few hours ahead, is crucial for a wide array of applications.  From individuals planning their daily activities, to farmers deciding whether to apply fertilizer, to meteorologists issuing timely warnings for severe weather events, accurate and timely nowcasts are essential.  Inaccurate precipitation forecasts can hinder disaster preparedness and response efforts, potentially leading to greater loss of life and property.  In fact, the WMO estimates that, over the past 50 years, 22\% of deaths and 57\% of economic losses caused by natural disasters were the result of “extreme precipitation” events \citep{Economist2023}.  However, nowcasting, particularly precipitation nowcasting, presents significant challenges, especially in tropical regions. 

In general, weather forecasting systems benefit greatly from availability of raw observations.  Doppler weather radars serve as the foundational instrumentation for the monitoring and forecasting of precipitation.  Their operational availability typically determines the precision and spatial resolution of meteorological forecasts within any given region.  However, coverage of ground-based weather radars is highly uneven across the globe.  While dense radar networks exist over North America, Europe and parts of East Asia, there is a severe lack of radar coverage in developing regions, oceans and largely uninhabited areas.  This further exacerbates the gaps in accuracy of precipitation forecasts between the Global North and the Global South (see Figure \ref{fig:csi_north_south}). 

Traditional Numerical Weather Prediction (NWP) methods play a significant, albeit evolving, role in precipitation nowcasting.  They serve as a cornerstone for understanding atmospheric dynamics and provide valuable context for shorter-term predictions.  However, they also have limitations when applied to the rapid timescales of nowcasting.  Running NWP models can be computationally expensive and time consuming, limiting their ability to produce frequent, low-latency updates needed for effective nowcasting \citep{Sun2014}.  For example, the High-Resolution Rapid Refresh (HRRR) model, produced by National Oceanic and Atmospheric Administration (NOAA), first collects and processes large amounts of observational data that feeds into their data assimilation system which runs on high-performance computing systems.  The initial conditions are then fed to the forecasting system also running on supercomputers to produce the forecasts.  This entire process takes about an hour and is limited to the CONUS region. 

Besides being more actionable in the near future, sub-hourly nowcasts are needed to capture the fine-scale details of convective precipitation which can develop and dissipate in under 30 minutes.  AI models promise lower latency, which could support forecasters in capturing these events in a way that is both accurate and timely.  While NWP methods have improved in spatial and temporal resolutions over the past few years, achieving a global forecast at a $0.05^{\circ} \times 0.05^{\circ}$ spatial resolution and 15-minute temporal resolution, with sub-hourly latency, remains a significant challenge for current global NWP systems.  The high-resolution forecast (HRES) from the European Centre for Medium-Range Weather Forecasts (ECMWF), while providing global coverage at a 9km resolution, is a medium-range model with a latency of several hours, making it unsuitable for the immediate, sub-hourly updates required for nowcasting.  Similarly, HRRR is a 3km spatial resolution model but available within the US only.  Additionally, NWPs continue to suffer from the problem of unequal skill in different parts of the world. 

The application of machine learning to medium-range weather forecasting has seen significant progress, with models like GraphCast  \citep{Lam2023}, GenCast \citep{Price2025}, NeuralGCM \citep{Kochkov2024}, Pangu-Weather \citep{Bi2023Pangu} and Fuxi \citep{Chen2023Fuxi} for medium-range forecasting demonstrating promising results.  This growing body of work, however, has not addressed the issue of accuracy gaps in different regions globally. Furthermore, the spatial and temporal resolutions of these models remain similar to their NWP counterparts, as these AI-based systems are built for an entirely different purpose than nowcasting.  Radar-based nowcasting methods using machine learning are able to overcome limitations of the traditional methods and showing considerable improvements in accuracy \citep{Piran2024,Ravuri2021,Espeholt2022}.  Although extremely effective in radar-rich parts of the world, they are inapplicable to most of the rest of the world due to radar-sparsity.  Satellite-based methods offer a potential solution, and some work has been done towards this, leveraging techniques such as optical flow are beginning to be adopted in data sparse regions, but have known limitations \citep{WMO2023}.  RainAI \citep{Sarabia2023} offers a method using EUMETSAT data as input and training against the OPERA network;  however, it is unclear whether that approach generalizes to regions without radar. \cite{Lebedev2019} propose a similar satellite-based approach training against the radars in Russia but mention the problem of overfitting to regions with radar and potentially risking coverage of other areas.
\begin{figure}[h!]
    \centering
    \includegraphics[width=0.9\textwidth]{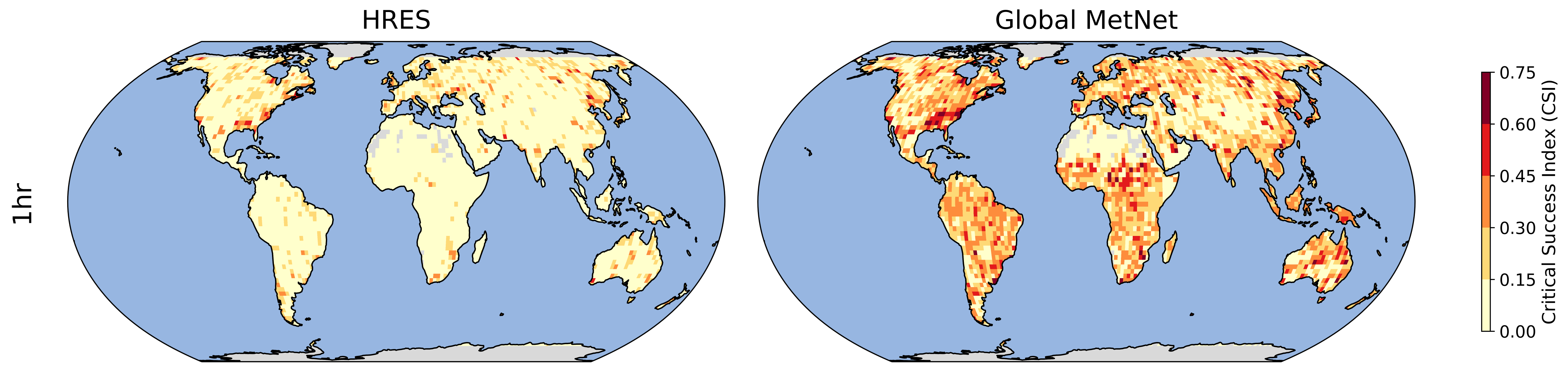}
    \caption{Critical Success Index (CSI) at a 1° resolution for the HRES and Global MetNet model at 1 hour lead time for 1.0 mm/hr of precipitation.}
    \label{fig:csi_main_map}
\end{figure}
This work presents a precipitation nowcasting model, Global MetNet, that is globally available but specifically designed to be highly performant in data sparse regions of the world. It bridges the accuracy gaps we see in the current state-of-the-art nowcasting models in most of the world where populations live (see Figure \ref{fig:csi_main_map}).  Extending our prior work on MetNet for regional nowcasting \citep{Espeholt2022}, this is a satellite observations based machine learning model with high spatial and temporal resolution that incorporates elements to make it easily operational.  Since ground radar is not available globally, our model leverages a global mesh of geostationary satellites as input, and to the best of our knowledge is the first system to use the Global Precipitation Mission's Combined Radar-Radiometer Precipitation algorithm dataset \citep{GPM_ATBD} as a training target.  The CORRA dataset combines data from a space-based dual-frequency precipitation radar with a microwave radiometer to create highly accurate estimates of rainfall. It provides near-global coverage and serves as a unique proxy for ground truth.  By leveraging this combination of observational data sources, our model provides nowcasts at a 15-minute resolution for the next 12 hours.  We evaluate our model against ground weather radar where available, calibrated and quality controlled rain gauges and the CORRA dataset where none of the other ground observations are available.  Our model outperforms industry-standard hourly forecasts globally, demonstrating its effectiveness in both data-rich and data-sparse regions.  We also show that an optimized HRES forecast, post-processed using our own ML model, is a stronger baseline than the raw HRES forecast itself.  Our work is especially critical in the tropics, where the lack of ground radar and other weather infrastructure limits the accuracy of the best-known current nowcasting methods. 

\section{Results}
\label{sec:results}

Here we present the results of the Global MetNet model compared to industry gold standard NWP forecasts, HRRR in the US, and HRES globally.  All results have been computed using the \cite{WeatherBenchX} framework. We compute metrics over various regions of the world because the varying climatologies can significantly impact the numbers.  We also show results for varying rates of precipitation from the category of light rain to heavy precipitation.  The results highlight substantial enhancements in predicting precipitation events across various lead times and geographical areas.  It is important to note that the results here take operational latencies into account.  For example, while HRES produces a nowcast for a 1-hour lead time, due to the operational latency, the forecast only becomes available after its valid time has already passed.  Hence, in the best-case scenario only the 7 hour lead time forecast of HRES is available as a 1 hour nowcast from any given initialization point (see Figure \ref{fig:timeline} in the supplement to help demonstrate). 

The Global MetNet model architecture has been designed to be flexible in the set of training datasets and we show results here for three different versions of our model with the only difference being the input datasets for training.  These model variations share the same model architecture but are trained independently allowing each one to optimize model parameters based on their respective inputs. The first model, called Global MetNet Nowcasting contains geostationary datasets and HRES NWP analysis and forecasts only as input.  To contrast this, we train a second model that includes high quality ground radar observations, called Global MetNet Nowcasting (with radar input).  Both of these models are trained with the following targets as separate output heads: the GPM CORRA dataset, ground radars from the US, Europe and Japan, and the GPM IMERG dataset (more in Table \ref{tab:training_targets} later).  A baseline model, called Global MetNet Post-processed HRES, is trained such that it takes only NWP data as input and trained to optimize the GPM CORRA dataset as target only.  This baseline model helps calibrate HRES against GPM CORRA dataset and makes for a much stronger baseline than the deterministic forecasts from HRES.  The primary goal of this baseline model is to show the importance of additional inputs other than NWP along with the strength of our model architecture.

We evaluate our forecasts against quality controlled ground radar datasets, which are considered the gold standard for precipitation measurements, and the GPM CORRA dataset to provide uniform global coverage. For all the following results, our test dataset spans one full year from June 2023 to May 2024. As a spaceborne satellite, the GPM CORRA dataset is not considered as high quality as ground radar  \citep{Speirs2017}, primarily because the GPM radar cannot see the precipitation all the way to the surface, and that it does not provide consistent global snapshots, with a revisit rate of 2.5 days; however, it makes for a uniform dataset to evaluate against globally, providing consistent coverage even over oceans, complex terrains or where radar is unavailable.  Note here that this dataset only captures sparse measurements and therefore a large enough validation dataset is required to be able to get less noisy evaluation against all possible precipitation rates.

\begin{figure}[h!]
    \centering
    \includegraphics[width=0.9\textwidth]{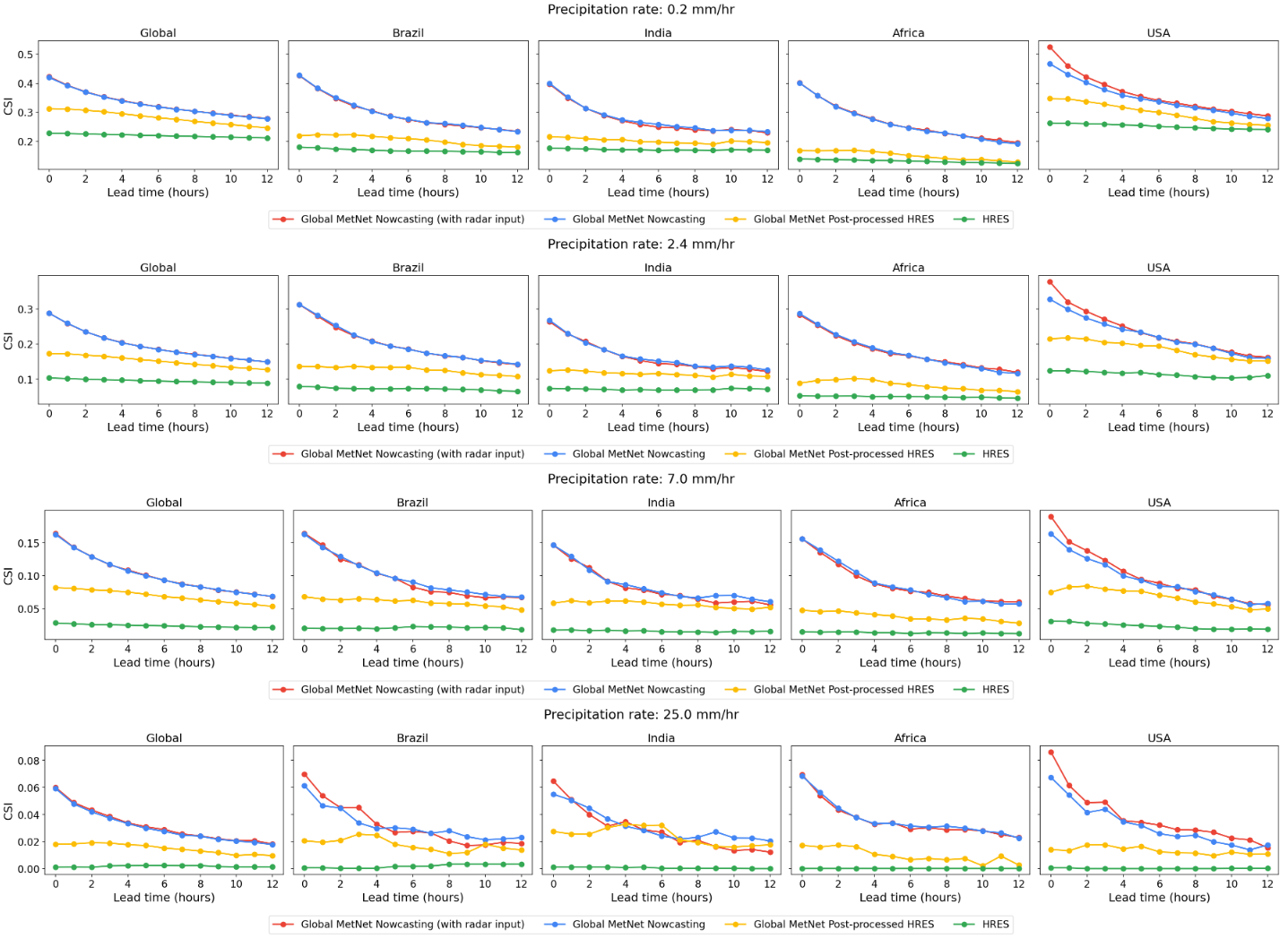}
    \caption{Critical Success Index (CSI) globally and for several regions (Brazil, India, Africa, and the USA), using the GPM CORRA dataset as ground truth at precipitation rates of 0.2 mm/hr (drizzle), 2.4 mm/hr (light rain), 7.0 mm/hr (heavy), and 25.0 mm/hr (very heavy). }
    \label{fig:csi_corra}
\end{figure}

Figure \ref{fig:csi_corra} shows results for our key metric, Critical Success Index (CSI). We see that globally and regionally, for all lead times and precipitation rates, Global MetNet continues to perform better than both the baselines HRES and post-processed HRES.  At $0.2\,\text{mm/hr}$ globally, MetNet shows a performance improvement of $\sim$0.18 CSI points over HRES for the first forecasting hour and narrows the gap between the performance of post-processed HRES at about 12 hours.  Even for higher precipitation rates of $25.0\,\text{mm/hr}$, MetNet performs much better where HRES is largely unable to predict these extreme events whereas post-processed HRES at least performs better than HRES. At that higher rate of precipitation, there is some visible noise in evaluation due to lack of sufficient observation data at these rates over any given region. Regionally, we see that the performance of HRES in the US is much higher than that over other regions, demonstrating the challenges with predicting chaotic precipitation in the tropics. Notably, the Global MetNet model trained with radar as an additional input, performs better only over regions where radar is included such as the USA. We do not see any influence of ground radar inputs in other places that do not have this data provided as an input to the model.

\begin{figure}[h!]
    \centering
    \includegraphics[width=0.9\textwidth]{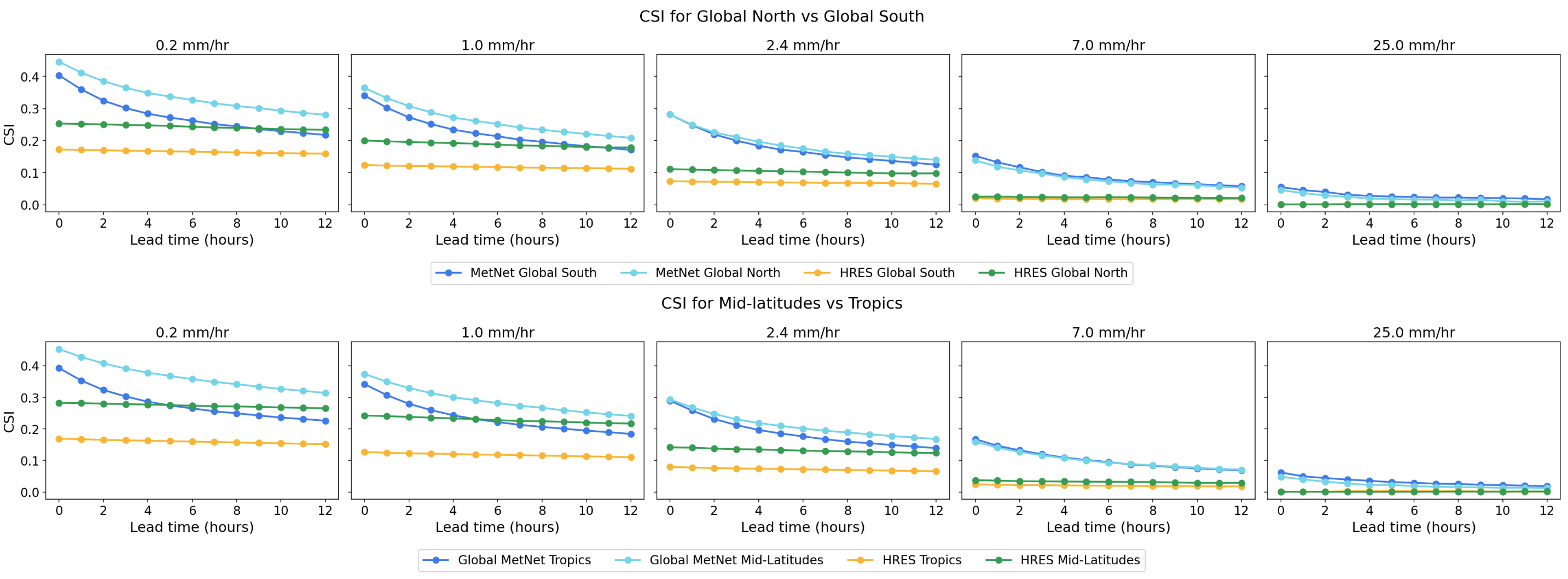}
    \caption{Forecasting Accuracy Gap: Critical Success Index (CSI) of Global MetNet vs. HRES in the Global South and Global North (top), and Tropics and Mid-Latitudes (bottom), validated against the GPM CORRA dataset at rates of $0.2, 1.0, 2.4, 7.0, \text{and } 25.0\,\text{mm/hr}$. Global North includes areas covering USA, Canada, Europe, Japan, and Australia. Global South includes regions covering India, South-east Asia, Middle-east, Africa, Brazil, Mexico, Central America and South America}
    \label{fig:csi_north_south}
\end{figure}

\begin{figure}[h!]
    % \centering
    % First subfigure
    \begin{subfigure}[b]{0.48\textwidth}
        \centering
        \includegraphics[width=\textwidth]{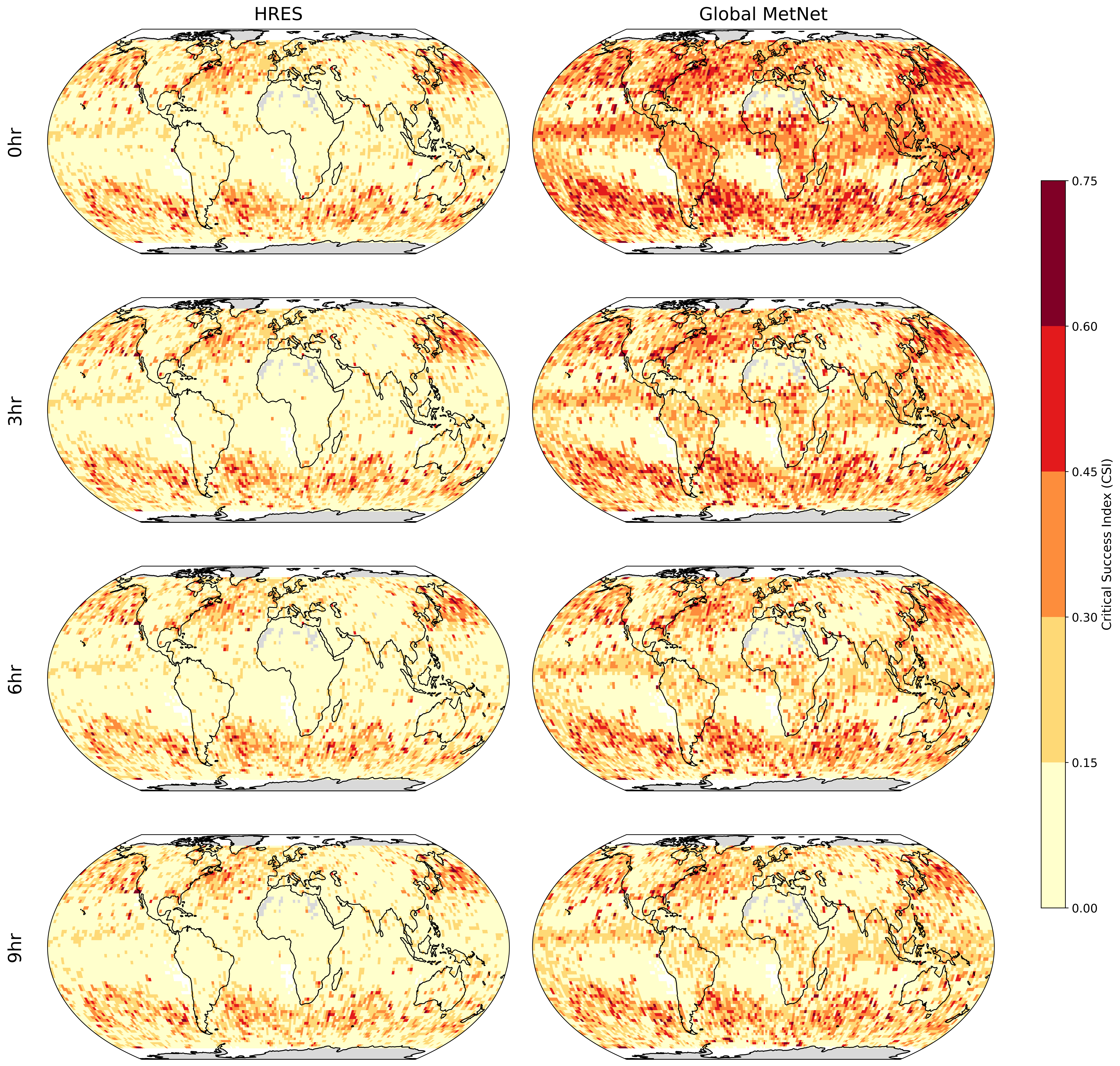}
        \caption{CSI for a precipitation rate of 1.0 mm/hr.}
        \label{fig:csi_1mm}
    \end{subfigure}
    \hfill % Adds horizontal space between the figures
    % Second subfigure
    \begin{subfigure}[b]{0.48\textwidth}
        \centering
        \includegraphics[width=\textwidth]{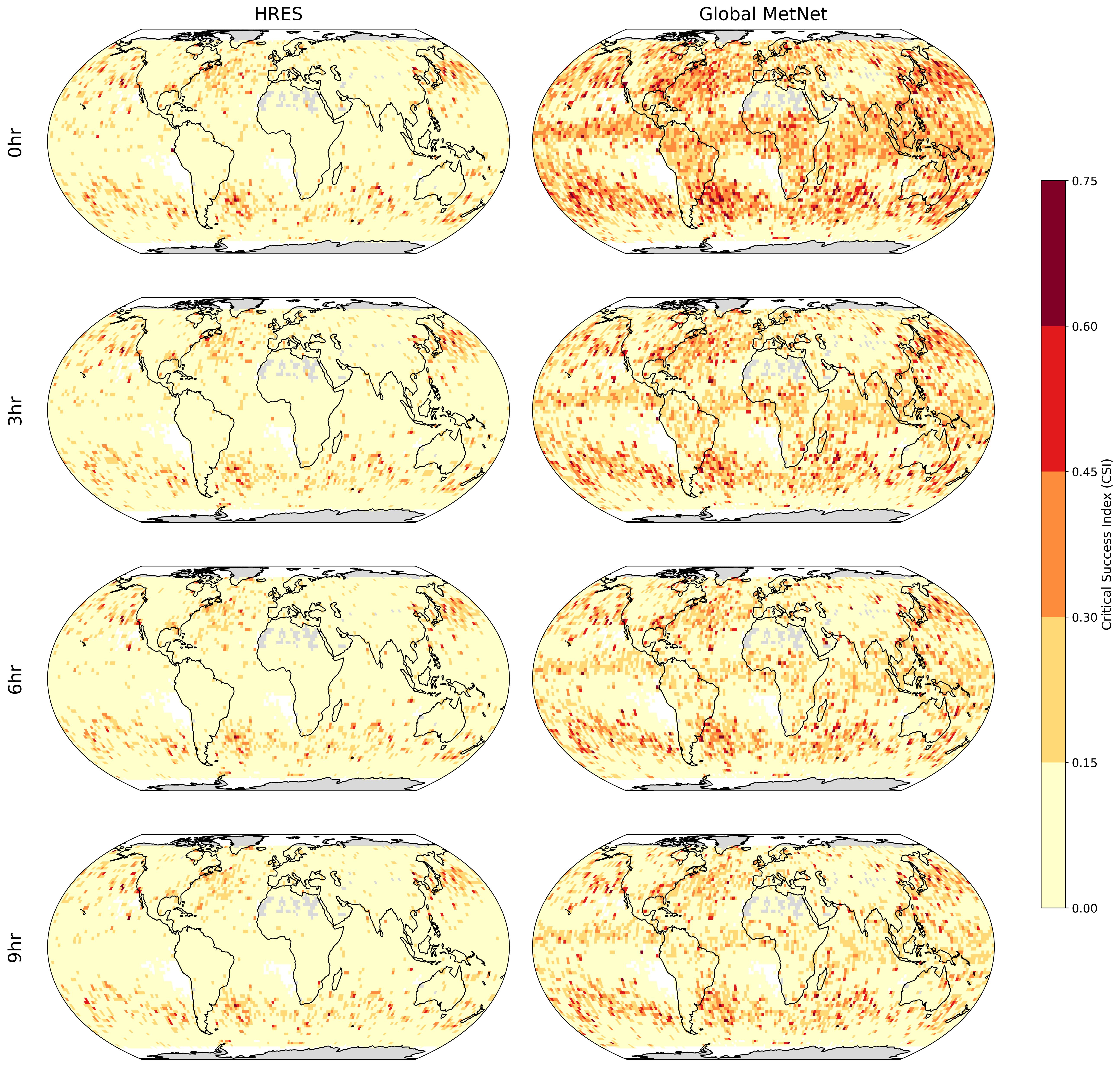}
        \caption{CSI for a precipitation rate of 2.4 mm/hr.}
        \label{fig:csi_2_4mm}
    \end{subfigure}
    \caption{Comparison of Critical Success Index (CSI) for HRES and Global MetNet nowcasts at different lead times (3, 6, 9, and 12 hours) for light (1.0 mm/hr) and moderate (2.4 mm/hr) precipitation.}
    \label{fig:csi_maps}
\end{figure}

Figure \ref{fig:csi_north_south} shows forecasting accuracy gap between the Global South and Global North and also between the tropics and the mid-latitudes. In Figure \ref{fig:csi_maps}, we plot the CSI scores for various regions on a map for better context in the improvements we see globally between HRES and Global MetNet. Remarkably, Global MetNet elevates the forecast skill in the Tropics and Global South (blue line) to a level that is comparable to, and for most lead times and precipitation rates exceeds the skill of the industry-standard HRES model in the data-rich Mid-latitudes and the Global North (green line). At $2.4\,\text{mm/hr}$ of precipitation, Global MetNet is able to close this forecasting accuracy gap. Overall, this doesn't just reduce the accuracy gap; it effectively eliminates the gap for certain conditions, representing a pivotal step toward global forecast equity.

\begin{figure}[h!]
    \centering
    \includegraphics[width=0.9\textwidth]{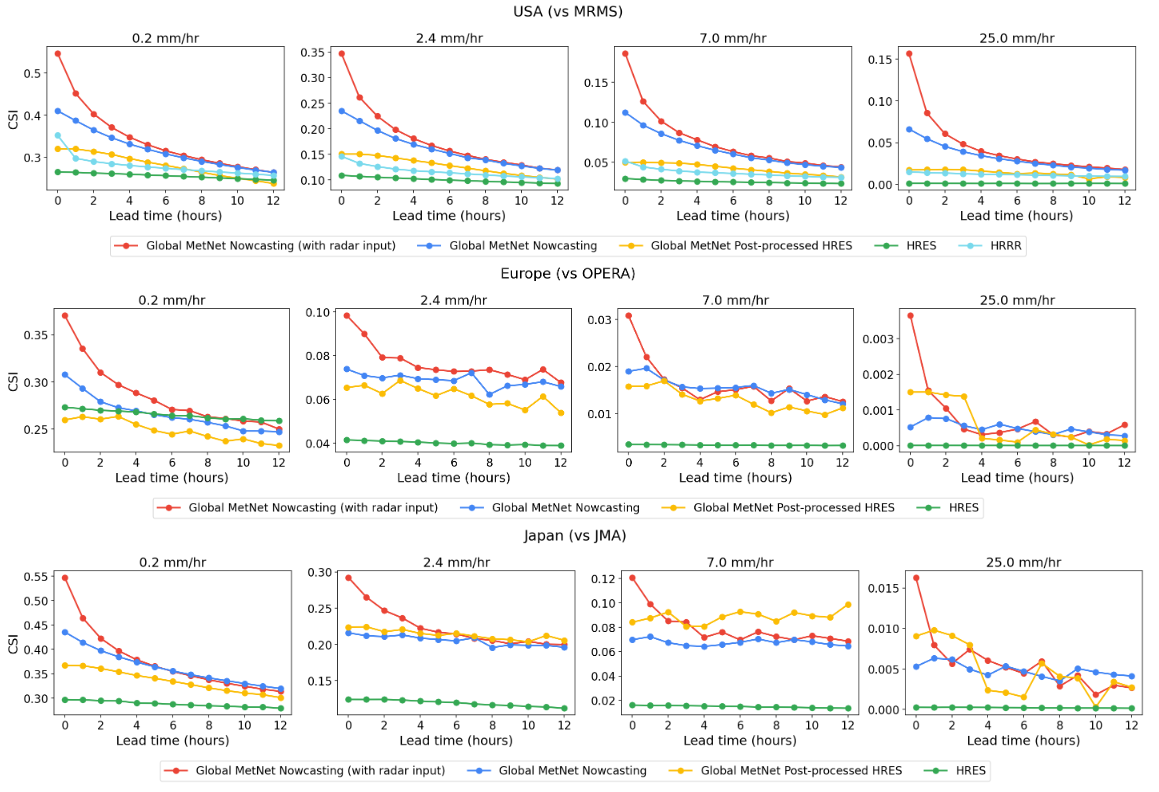}
    \caption{Critical Success Index (CSI) for Global MetNet models vs. NWP baselines in the US (vs. MRMS), Europe (vs. Opera), and Japan (vs. JMA) at precipitation rates of 0.2, 2.4, 7.0, and 25.0 mm/hr. }
    \label{fig:csi_radar}
\end{figure}

Next, in Figure \ref{fig:csi_radar}, we present results evaluated against ground radar based precipitation estimates over the US from \cite{MRMS}, over Europe from the OPERA network \citep{Opera2014}, and over Japan from the \cite{JMA_Radar} radars. We can see that the Global MetNet model, even when trained without high quality ground radars, outperforms global and regional NWP HRRR at all lead times up to 12 hours and at all rain rates. The performance of the model trained with the regional radars as an input is the highest up to 6 hours of lead time at all precipition rates. Note here that the prediction of Global MetNet models is optimized for the GPM CORRA dataset, whereas we evaluate against radars in this figure and hence, there is some loss inherently due to the discrepancy in observations between GPM CORRA and radar datasets. At higher rates, such as 25 mm/hr, some noise is visible due to lack of sufficient observation data at those points. These results demonstrate the high skill of the model against the best available ground truth even when the 'gold standard' of ground-based radar networks are not available during training or inference.  Achieving good skill despite the absence of radar inputs is particularly critical in the Global South where radars are not widely available. This indicates the model is learning meteorologically sound patterns, rather than simply overfitting to the characteristics of a single sensor type.

\begin{figure}[h!]
    \centering
    \includegraphics[width=0.9\textwidth]{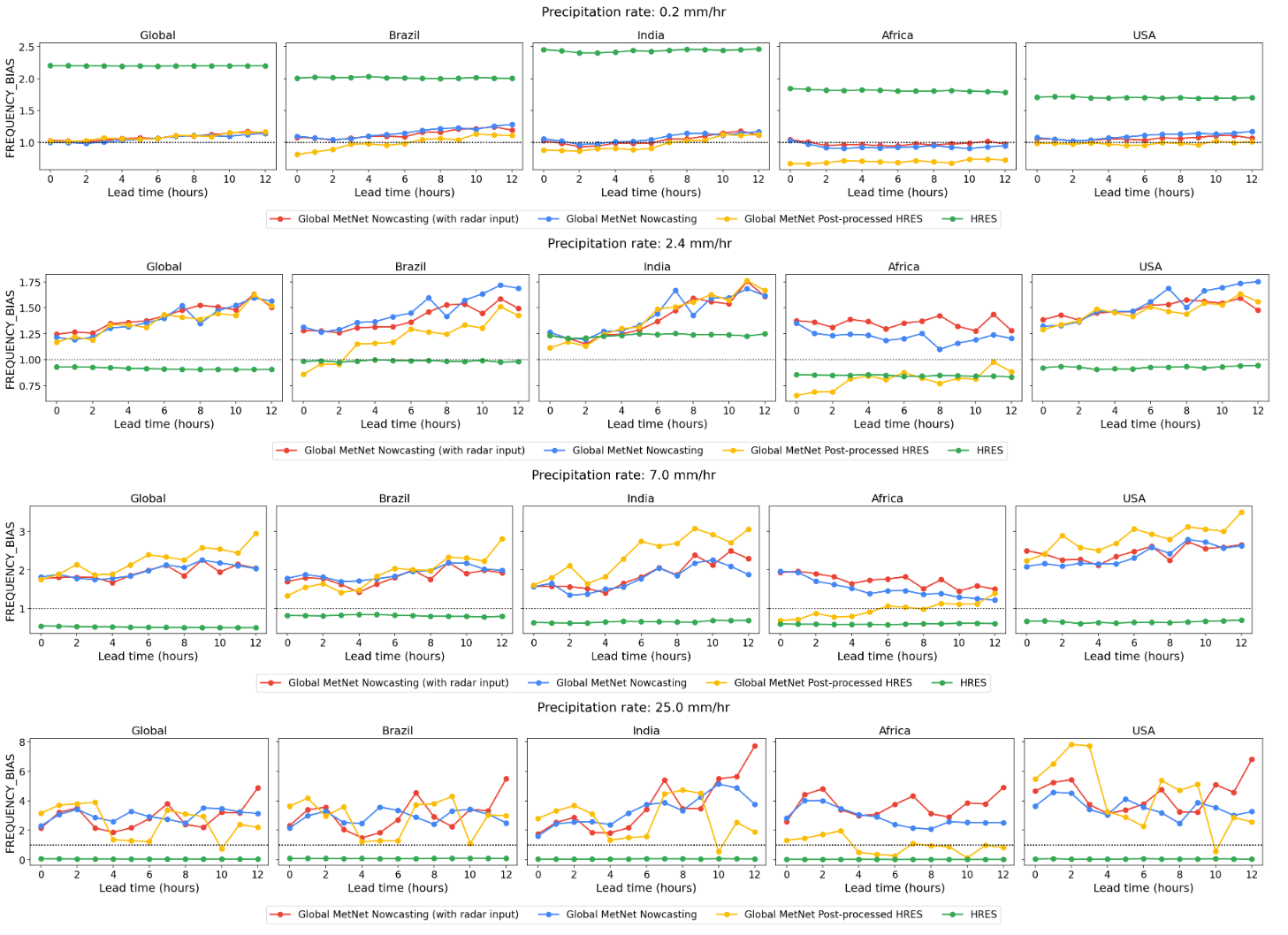}
    \caption{Frequency Bias Globally and by Region for Precipitation Rates of 0.2, 2.4, and 25.0 mm/hr. }
    \label{fig:freq_bias_corra}
\end{figure}

When looking at the frequency bias of the Global MetNet models compared to HRES, in Figure \ref{fig:freq_bias_corra}, we note that there is some variation in the bias at varying lead times, rates of precipitation and regionally as well.  For the 0.2 mm/hr precipitation rate, we see that Global MetNet’s bias stays close to 1 at all lead times both globally and regionally, whereas raw HRES tends to overpredict these lower thresholds more than twice.  As we get to the higher rates, we can see that Global MetNet and post-processing HRES leads to an overprediction whereas HRES underpredicts globally.  It should be noted that for more extreme precipitation it is better to over-predict and issue sufficient warning to end-users rather than leave them unprepared, this is commonly known as wet bias.  As uncertainty of the forecast increases with lead time for higher precipitation rates, Global MetNet tends to overpredict accordingly.  It is important to note here that the probabilistic inference from Global MetNet is categorized by applying probability thresholds optimizing for the CSI metric, which results in sub-optimal frequency bias scores.  However, if one was interested in specifically optimizing frequency bias then it is possible to apply thresholds to optimize that instead and we noticed that it does not decrease the performance of CSI much at all. 

\begin{figure}[h!]
    \centering
    \includegraphics[width=0.9\textwidth]{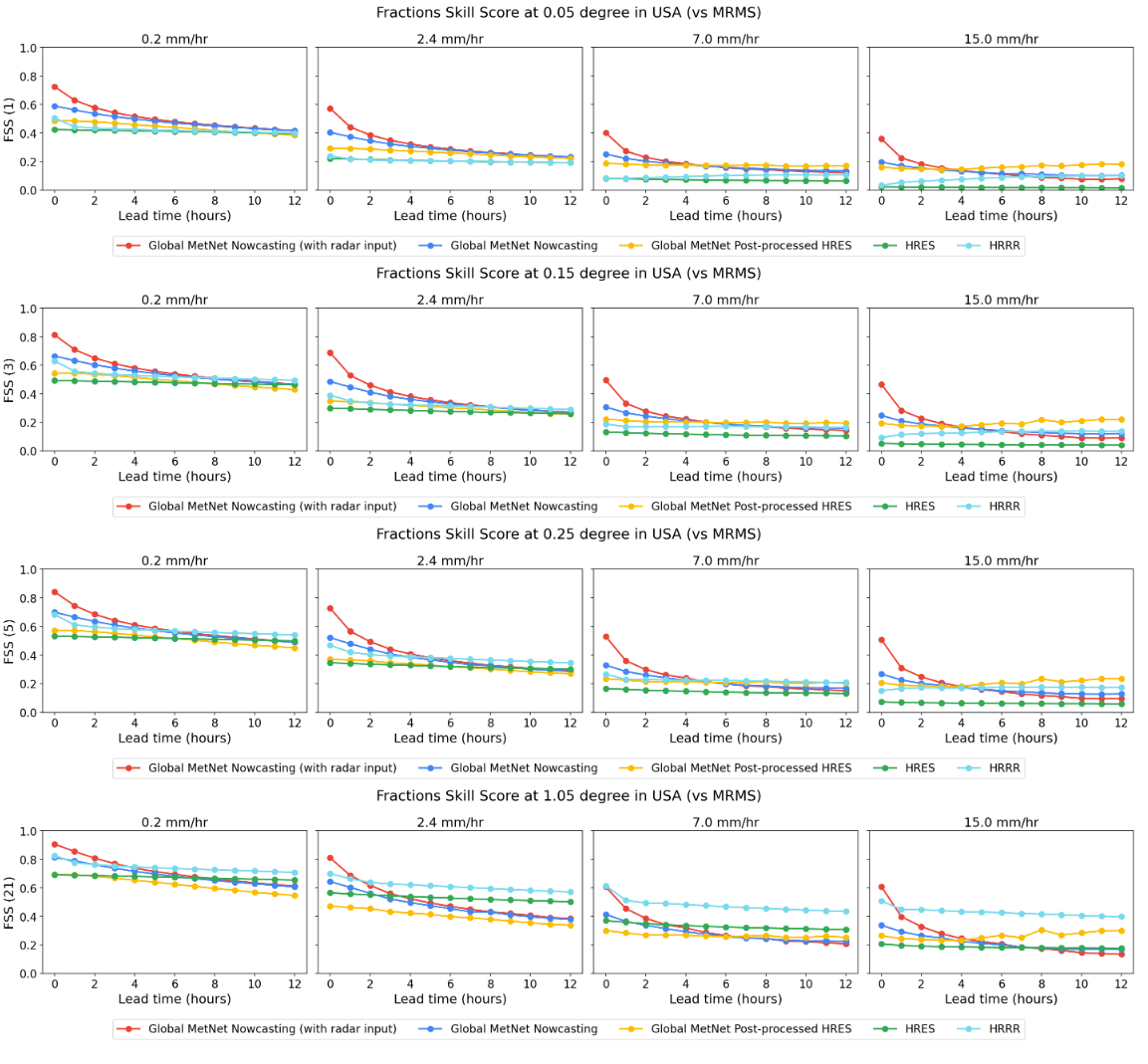}
    \caption{Fractions Skill Score (FSS) of Global MetNet vs. NWP Baselines in the US (vs. MRMS) for Various Precipitation Rates (0.2, 2.4, 7.0, and 25.0 mm/hr) across a Range of Spatial Neighborhoods (0.05° (FSS(1)) to 1° (FSS(21)).}
    \label{fig:fss_usa}
\end{figure}

We also show results for a spatial verification metric, fractions skill scores (FSS)\citep{Roberts2008} for varying sizes of pixel neighborhoods from 0.05$^{\circ}$ to 1$^{\circ}$.  In Figure \ref{fig:fss_usa}, we show results of the Global MetNet models vs NWP models (HRES and HRRR) in the US using MRMS as the ground truth.  Due to the narrow swaths of the GPM CORRA dataset it is not possible to apply spatial verification metrics such as FSS at much coarser resolutions therefore we provide results here against a dense ground truth like MRMS.  The FSS quantifies the ability of a forecast to correctly identify precipitation patterns at different spatial scales, with higher values indicating better skill.  Fractions skill score is also an important metric to look at that avoids the double penalty problem \citep{ECMWF_Verification} that metrics like CSI may suffer from placing NWP models at a disadvantage.  Overall, Global MetNet has higher skill than both the other baselines at all of these neighborhood sizes, precipitation rates and at all lead times.

As expected, looking at Figure \ref{fig:fss_usa}, we note that the FSS generally decreases as the neighborhood size decreases (from 1° to 0.05°).  This reflects the increasing difficulty of accurately predicting fine-scale precipitation features at higher resolution.  MetNet is able to capture even the more chaotic, heavier precipitation events also more skillfully than NWP models at earlier lead times and meets the HRRR model by hour 12 at finer resolutions.  While HRRR shows higher skill at an extremely coarse 1° neighborhood, this primarily reflects its ability to correctly place a large weather system within a very large general area. For the high-resolution scales that are most meaningful for nowcasting applications (e.g., 0.05° to 0.25°), Global MetNet consistently demonstrates superior skill in capturing the actual location and spatial structure of precipitation, making it a more valuable tool for localized warnings.

\section{Global MetNet}
\label{sec:methods}

\subsection{Datasets}
\label{subsec:datasets}
This section outlines the multi-modal datasets used by Global MetNet, distinguishing between non-time-sensitive training targets and low-latency input features required for real-time inference.  These datasets vary in spatial and temporal scales and real-time latencies, collectively enabling global coverage and enhanced prediction capabilities.  Further details on each dataset are available in the supplement. 

\subsubsection{Training Targets}
\label{ssubsec:training_targets}
An ML model is optimized by taking in a set of inputs and corresponding targets to train against.  Hence, during inference when the model is operationalized, the datasets used as model training targets do not need to be available with a low latency.  This gives us an opportunity to use calibrated observations in our model as training targets. Ideally, a global network of ground-based weather radars would provide the highest quality, high-resolution precipitation data for training. However, in reality, this is a challenging task for a number of reasons.  Radars can be expensive to install and maintain such as over the ocean or mountains or in places lacking relevant infrastructure and trained personnel.  Many times, even if radars exist they are owned by city governments or by different organisations even within a country, and their data is not easily available for use by external organisations.  Furthermore, even if the raw radar data is readily available for use it can be noisy picking up false signals from flocks of birds, wind farms and sun interference.  A mountainous terrain or presence of tall buildings close to the station can further lead to inaccurate data.  This raw radar data requires significant processing and cleanup before it can be used as a training target or for validation. 

\begin{figure}[h!]
    \centering
    \begin{subfigure}[b]{0.50\textwidth}
        \centering
        \includegraphics[width=\textwidth]{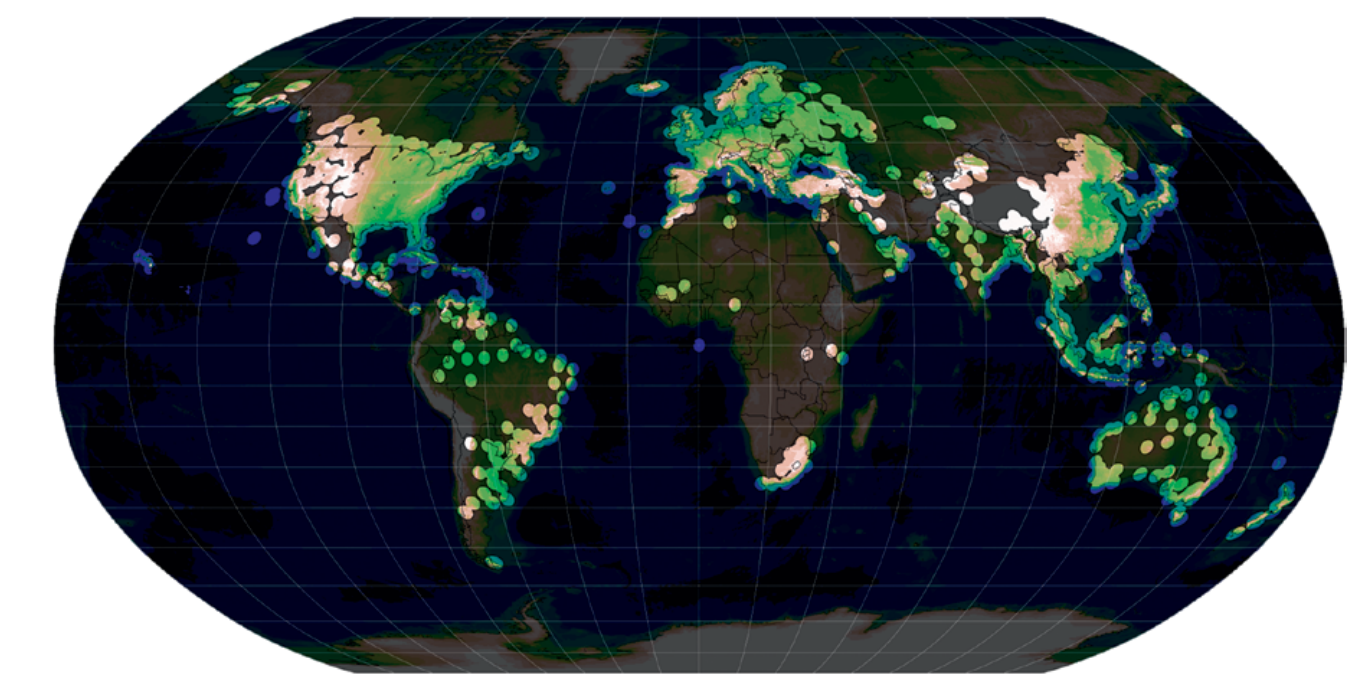}
        \caption{Radar coverage globally \citep{Saltikoff19}. This figure is for illustration purposes as noted by Saltikoff et al. in Figure 1 of their paper and may not be up-to-date.}
        \label{fig:radar_coverage}
    \end{subfigure}
    \hfill % This command adds horizontal space between the figures
    \begin{subfigure}[b]{0.46\textwidth}
        \centering
        \includegraphics[width=\textwidth]{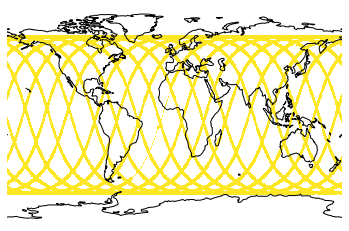}
        \caption{15 consecutive orbital swaths sampled by GPM CORRA, demonstrating the spatial footprint observed by the satellite over the course of a full day.}
        \label{fig:gpm_swath}
    \end{subfigure}
    \caption{Global data coverage maps for the training targets.}
    \label{fig:coverage_maps}
\end{figure}

To facilitate validation and training of the model on precipitation measurements from other parts of the world and especially the tropics, we make use of NASA’s Global Precipitation Measurement (GPM) mission’s dual-frequency precipitation radar satellite.  GPM provides a precipitation estimate using the CORRA algorithm, which is sparse but provides global coverage (see Figure \ref{fig:coverage_maps} for a map of global coverage).  Additionally, we use the \cite{NCAR_GPM_Guide} final precipitation estimate as another training target, which is dense, but has potential inaccuracies.  Table \ref{tab:training_targets}, summarizes the features of the training targets used by the Global MetNet model, where the target type shows that the GPM CORRA data is the main target which makes the actual predictions used in all of our evaluations and results. The other datasets serve as auxiliary training targets.

\begin{table}[h!]
    \centering
    \caption{This table summarizes the training targets and their properties.}
    \label{tab:training_targets}
    % Use tabularx and set its total width to the text width
    % The 'X' column type allows text to wrap automatically
    \begin{tabularx}{\textwidth}{X l l X l} 
        \toprule
        \textbf{Dataset} & \textbf{Spatial Resolution} & \textbf{Target Patch Size} & \textbf{Coverage} & \textbf{Target Type} \\
        \midrule
        GPM CORRA & $0.05^{\circ} \times 0.05^{\circ}$ & $3600 \times 7200$ & Sparsely global & Main \\
        Ground Radars & $0.05^{\circ} \times 0.05^{\circ}$ & $3600 \times 7200$ & Dense in US, Europe, Japan & Auxiliary \\
        IMERG Final & $0.1^{\circ} \times 0.1^{\circ}$ & $1800 \times 3600$ & Dense globally & Auxiliary \\
        \bottomrule
    \end{tabularx}
\end{table}

\subsubsection{Training Input Features}
\label{ssubsec:training_inputs}
Unlike the training targets that do not need to be available in real-time during operations, the training features should be available at least within a few hours of inference time to be relevant to the predictions. Table \ref{tab:input_datasets} outlines the datasets we use as gridded inputs along with their real-time latencies. These latencies are baked into the training dataset by fetching older data (corresponding to its real-time latency) than the one at the initialization time of each training example. In the table, we also mention how many historical timestamps we use for each of the inputs.

\begin{table}[h!]
    \centering
    \caption{Input Datasets. }
    \label{tab:input_datasets}
    \resizebox{\textwidth}{!}{%
    \begin{tabular}{lcccc}
        \toprule
        \textbf{Dataset} & \textbf{Spatial Resolution} & \textbf{Real-time Latency} & \textbf{\# channels} & \textbf{\# historical timestamps} \\
        \midrule
        Ground Radars & $0.05^{\circ} \times 0.05^{\circ}$ & $\sim$30 mins & 1 & 6 timestamps, 15 mins apart \\
        Geostationary Satellite Mosaics & $0.05^{\circ} \times 0.05^{\circ}$ & $\sim$60 mins & 17 & 3 timestamps, 30 mins apart \\
        HRES atmospheric variables & $0.1^{\circ} \times 0.1^{\circ}$ & 6 to 12 hours & 63 & 1 last available timestamp \\
        HRES surface variables & $0.1^{\circ} \times 0.1^{\circ}$ & 6 to 12 hours & 40 & 1 last available timestamp \\
        IMERG Early & $0.1^{\circ} \times 0.1^{\circ}$ & 5 to 6 hours & 1 & 6 timestamps, 30 mins apart \\
        Elevation & $0.05^{\circ} \times 0.05^{\circ}$ & - & 1 & N / A \\
        Latitude - Longitude & $0.05^{\circ} \times 0.05^{\circ}$ & - & 2 & N / A \\
        \bottomrule
    \end{tabular}
    }
\end{table}

The geostationary satellite mosaics is a special dataset that we create through blending and calibration of multiple satellites and we go into the details of it next. Information on the rest of the inputs can be found in Supplement ~\ref{supp:datasets}.

\subsubsection{Geostationary Mosaics}
We use a total of 7 geostationary satellites as inputs to our model, that are combined into a mosaic to provide global coverage. Table \ref{tab:satellites} outlines the coverage provided by each of the satellites and the agencies that maintain them.  We also use equivalent older satellites for model training, where available. We use the satpy library to do the parsing and reprojecting of the raw data into 18 mosaics at varying wavelengths from 0.47 \textmu m to 13.3 \textmu m. Supplement \ref{supp:geosat} provides more details on each band in the mosaic. 

\paragraph{Mosaic Time Resolution and Latency}
We make mosaics at 30 minute intervals.  This is the lowest common multiple for any dataset that contains Meteosat Second Generation satellites.  Our data delivery delays are about half an hour and our processing time is under half an hour.  This means our realtime mosaics lag actual real time by about an hour in total. 

\paragraph{Blending and Calibration}
We use level 1b calibrated data for our mosaics.  This gets all the data in reflectance units for the visual bands and brightness temperature (Kelvin) for the IR bands.  We also use a Gaussian center weighted blended average to blend the data together.  This avoids any artifacts at the boundaries of the different satellite coverage areas.  We try to blend each mosaic at the highest resolution available for any input to the given mosaic, but we cap resolutions to 1km nominal pixel size for runtime reasons.

\begin{table}[h!]
    \centering
    \caption{Geostationary satellites used for creating mosaics. }
    \label{tab:satellites}
    \begin{tabular}{lll}
        \toprule
        \textbf{Satellite Name} & \textbf{Region Covered} & \textbf{Agency} \\
        \midrule
        Meteosat-11 & Europe/North Africa & EUMETSAT \\
        Meteosat-9 & Indian Ocean & EUMETSAT \\
        Meteosat-12 & Europe/North Africa & EUMETSAT \\
        Himawari-9 & East Asia \& Western Pacific & Japan Meteorological Agency \\
        GOES-19 & Eastern Americas \& Atlantic Ocean & NOAA \\
        GOES-18 & Western Americas \& Pacific Ocean & NOAA \\
        GK-2A & East Asia \& Western Pacific & Korea Meteorological Administration \\
        \bottomrule
    \end{tabular}
\end{table}

\subsection{Model Setup}
\label{subsec:model_setup}
This section details the data processing steps, model architecture, and the approach to generating probabilistic outputs. 

\subsubsection{Dataset Processing}
\label{ssubsec:dataset_processing}
The datasets were split into separate partitions for model development and evaluation.  The development dataset spans from 2018 to 2023 that we further split into a dataset for training the ML model and parameter optimization (January 1, 2018, to April 30, 2022), and a smaller held-out set for fitting the probability thresholds (May 15, 2022, to May 15, 2023).  Finally, the test dataset, covering the period from June 1, 2023, to May 31, 2024, was designated for final model evaluation and performance assessment. 

Before training, all datasets were preprocessed for consistency and quality.  All the datasets, except for the NWP data, were resampled to a consistent $0.05^{\circ} \times 0.05^{\circ}$ spatial resolution.  All the $0.05^{\circ} \times 0.05^{\circ}$ datasets undergo a space-to-depth \citep{Wang2020} operation with a block size of 2 which stacks each block of pixels to create more channels, which allows the model to analyze spatial patterns at different scales more efficiently.  The NWP data, on the other hand, was resampled to a $0.1^{\circ} \times 0.1^{\circ}$ resolution and no space-to-depth operation is applied to it.  Space-to-depth operation on higher resolution datasets was necessary firstly, to fit the data into the memory constraints and secondly, allowing concatenation of these higher resolution datasets with the lower resolution NWP data. This processing step brought all input datasets to a consistent effective grid size of $1800\times3600$ pixels before being fed into the model. 

We then normalize all of the input datasets to a zero mean and unit standard deviation values.  The precipitation inputs from radar sources are normalized using log normalization due to the high skew of precipitation data.  We then handle the missing or invalid data by replacing it with $0$s.  We also append each of the input datasets with timedeltas from the initialization time to inform the model.  These timedeltas were effectively added as extra channels. All the time slices of the inputs are concatenated along the channel dimension, then all the inputs are also concatenated together along the channel dimension to produce the final inputs to the model. 

Since the global data is represented through a rectangle we add a context of 18 degrees on each left and right edges of this rectangle to avoid any artificial border artifacts.  This brings the entire input data to a spatial dimension of $2160 \times 3600$. 

Instead of using a recurrent layer like an LSTM to process the time sequence of inputs, we concatenate the features from different input timesteps along the channel dimension.  This creates a very wide tensor that the subsequent convolutional layers will process.  This is a simpler but potentially effective way to provide temporal context. 

For the training data, target patches containing only missing values for any given lead time were mostly excluded and only a small percentage of such samples were kept chosen at random.  We had to do this as the GPM CORRA data is quite sparse and very many target lead times only contained missing values.  This ensures the model learns from valid precipitation data and prevents it from being trained on patches with no information.  By filtering out these entirely empty patches, the model's training is focused on meaningful precipitation patterns and values.  The targets are discretized by 30 different precipitation rates and any precipitation rate that is beyond a reasonable range of 2 meters/hour is replaced with a value of 0. 

\subsubsection{Model Architecture}
\label{ssubsec:model_arch}
At its core, Global MetNet, like its predecessors MetNet and MetNet-2 use an encoder-decoder structure.  The encoder processes the preprocessed input tensor, learning a compressed representation of current and past weather conditions.  The decoder takes this learned representation and generates forecasts at future lead times for various training targets configured as output heads. Here are some of the key architectural features:
\begin{itemize}
    \item \textbf{Conditioning with Lead Time:} Similar to MetNet-2, we encode the lead time as a one-hot embedding with indices from 0 to 721 representing the range between 0 and 12 hours with a 15 min interval and map them into a continuous 32-dimensional representation.  Instead of feeding the lead time embedding as an input, the embedding is applied both as an additive and multiplicative factor \citep{Perez2018} to the model inputs and to hidden representations before each activation function.  This ensures that the internal computation in the network depends directly on lead time. 
    \item \textbf{Initial Downsampling:} The concatenated input features are first passed through another space\_to\_depth operation.  This further reduces spatial resolution and increases channel depth, preparing the data for the main convolutional stack. 
    \item \textbf{Deep Residual Network:} The core of the encoder is a stack of residual blocks.  Residual connections help in training very deep networks by allowing gradients to flow more easily. 
    \item \textbf{Multiple Stages:} The encoder has 4 stages of these residual blocks. 
    \item \textbf{Number of Blocks per Stage:} Each stage consists of 8 residual blocks. 
    \item \textbf{Channels per Stage:} The number of feature channels increases from 256 in the first stage to 384 in the subsequent stages.  This allows the network to learn increasingly complex features. 
    \item \textbf{Cropping:} After each stage of residual blocks, a cropping operation is applied.  This progressively reduces the spatial extent of the feature maps. This is done because as network depth and neuron receptive fields increase, border information becomes less relevant for predicting the central area. 
    \item \textbf{Upsampling and Final Convolution:} After the final residual blocks and cropping, features are upsampled by repeating values to their initial resolution before passing through a final convolutional layer. Heads that require a higher output resolution than the encoder receive further upsampling and convolutional layers. 
\end{itemize}

\subsubsection{Training and Optimization Features}
\label{ssubsec:training_opt}
\begin{itemize}
    \item \textbf{Data Type:} The training casts all input data to bfloat16 for faster training and reduced memory usage with minimal precision loss on TPUs. 
    \item \textbf{Optimizer:} Uses the Adam optimizer with an initial learning rate of 3e-4 with a step change mid way through training at a lower rate of 1.5e-4. 
    \item \textbf{Polyak Averaging:} Averages model weights over training steps, which can lead to better generalization. 
    \item \textbf{Memory Optimization:} Enables gradient checkpointing (rematerialization) for input preparation, ResNet blocks, and heads.  This saves memory by recomputing activations during the backward pass instead of storing them all, crucial for large models. 
    \item \textbf{Hardware Configuration:} The training job is executed on a 16x16 Dragonfish TPU pod, which effectively has 256 TPU chips and 512 TPU cores in total. 
\end{itemize}

\subsubsection{Probabilistic Output Heads}
\label{ssubsec:prob_outputs}
The model uses multiple output 'heads,' each optimized for a specific prediction target, resolution, and lead time.  This allows each head to be optimized for the specific characteristics of its target variable while sharing the core of the encoder weights. 

In contrast to NWPs, that model uncertainty with ensemble forecasts, Global MetNet outputs a marginal probability distribution for precipitation at each location using a full categorical Softmax. Thus, each output head is discretized into bins and the model outputs the probability of precipitation for each bin for each lead time. This probabilistic approach enables a more comprehensive assessment of forecast uncertainty and improves the practical utility of the nowcasts for decision-making. 

Once the model has finished training on the training split of the dataset, we compute optimal probability thresholds for each discrete bin and each lead time. These thresholds are found by maximizing the CSI score on a held-out evaluation dataset.  The probability thresholds, a value between 0 and 1, that results in the highest CSI on aggregate on this evaluation dataset gets fixed for future inferences and final metrics computation on the testing dataset. 

\section{Case Studies and Evaluation}
\label{sec:case_studies}

To assess Global MetNet's effectiveness in real-world scenarios, this section presents case studies focusing on high-impact precipitation events. A crucial aspect of this evaluation is accounting for the significant differences in operational latency between the models. HRES forecasts have a latency of approximately six hours, whereas Global MetNet generates forecasts in under a minute. To ensure a fair and operationally relevant comparison, our analysis visualizes the earliest available forecast from each model for a given point in time, as illustrated in Figure~\ref{fig:timeline}. For these comparative visualizations:
\setlength{\topsep}{0pt}
\begin{itemize}[topsep=0pt, partopsep=0pt]
    \item HRES is represented by its direct, deterministic forecast value.
    \item Global MetNet's visualization is derived from its probabilistic output. The model predicts probabilities for several precipitation rates ($0.2, 1.0, 2.4, 5.0, 7.0, 10.0, 15.0,$ and $25.0$ mm/hr). These probabilities are then converted into a single deterministic forecast by applying thresholds that are optimized to maximize the Critical Success Index (CSI), a process detailed in Section~\ref{ssubsec:prob_outputs}. The highest precipitation rate identified through this process is then displayed.
    \item IMERG Final serves as an observational benchmark to estimate actual precipitation during the event.
\end{itemize}

\subsection{Deep Convective Systems}
\begin{figure}[h!]
    \centering
    \begin{subfigure}[b]{0.54\textwidth}
        \centering
        \includegraphics[width=\textwidth]{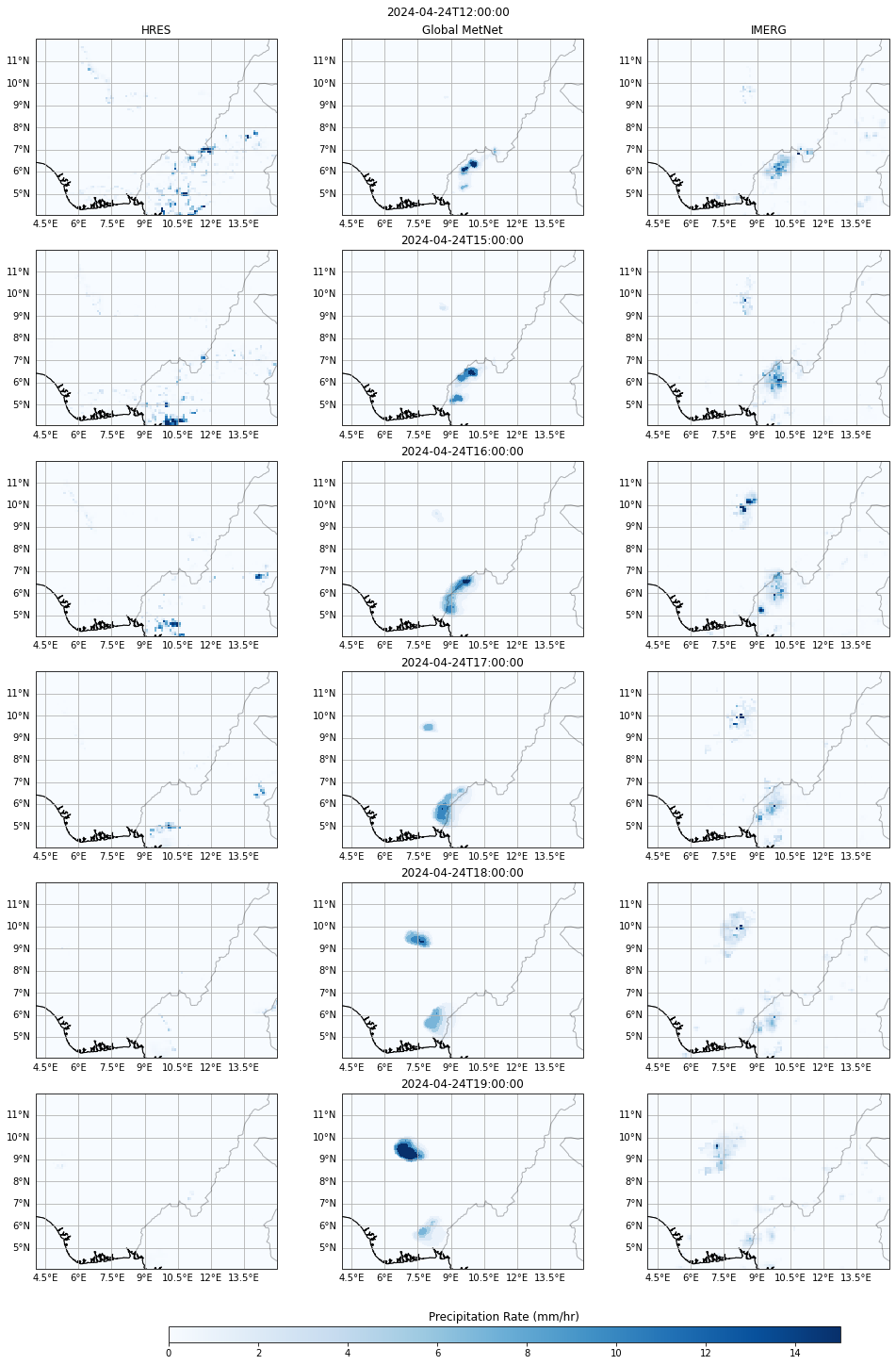}
        \caption{Forecast visualized for HRES (left), Global MetNet (middle) and IMERG (right).}
        \label{fig:MCS_WA}
    \end{subfigure}
    \hspace{1em} % Adds horizontal space between the figures
    \begin{subfigure}[b]{0.33\textwidth}
        \centering
        \includegraphics[width=\textwidth]{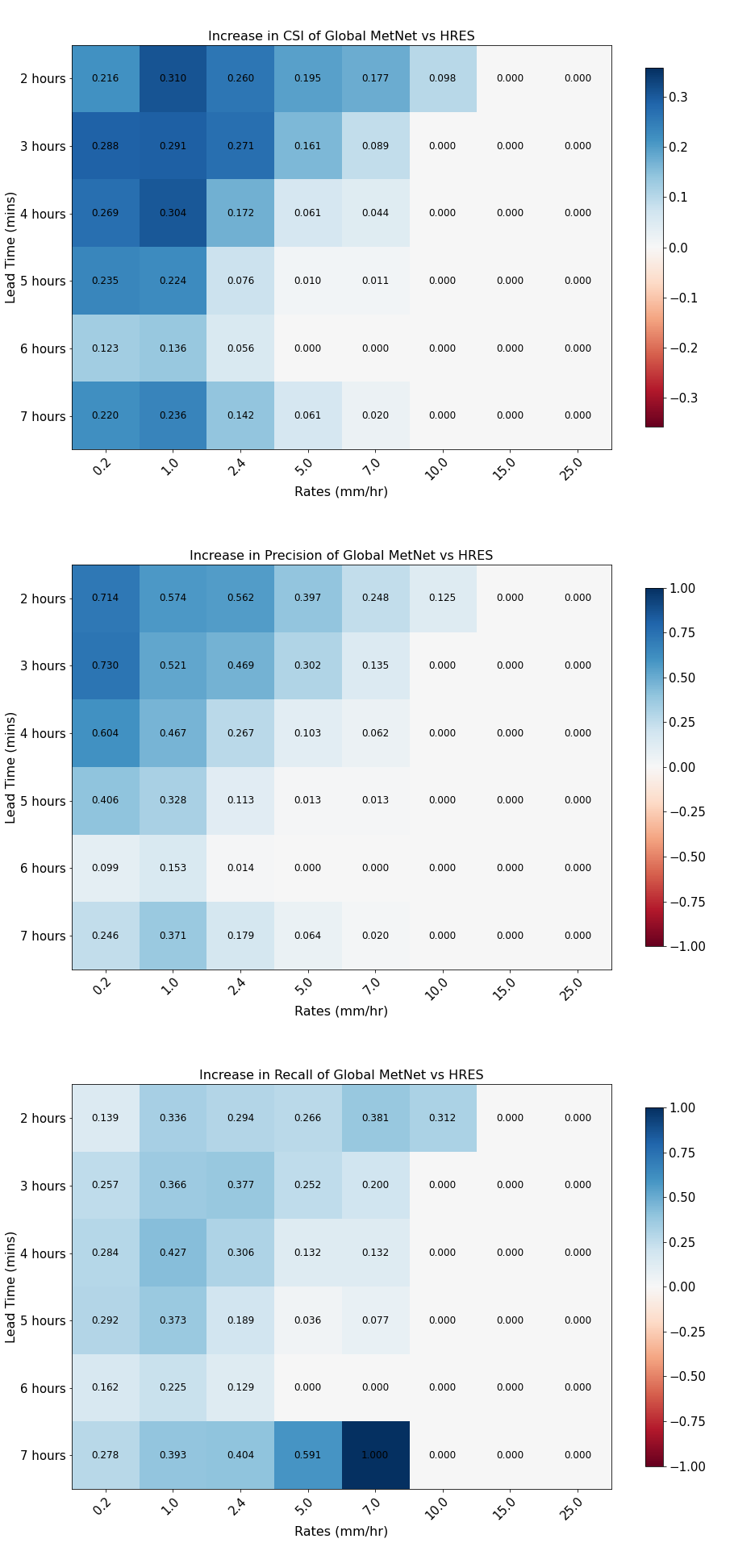}
        \caption{The difference in CSI, Precision and Recall  of the Global MetNet forecast over HRES forecast per lead time and per rate. Positive values imply an increase in metric and negative values imply a decrease.}
        \label{fig:MCS_WA_stats}
    \end{subfigure}
    \caption{Case study of a deep convective system in West Africa showing nowcasts available at 12:00 UTC, April 24, 2024 leading up to 19:00 UTC.}
    \label{fig:mcs}
\end{figure}

Figure \ref{fig:mcs} displays a side-by-side comparison of the HRES and Global MetNet forecasts against IMERG satellite precipitation estimates for a deep convective system that developed in West Africa on April 24, 2024.  The series of forecasts visualizes the models' performance in capturing the thunderstorm's development from 12:00 UTC up until 19:00 UTC. The HRES model is initialized at 06:00 UTC and the Global MetNet model is initialized at 11:58 UTC, allowing availability of a forecast from both the models for 12:00 UTC. The near-complete absence of the system in the HRES forecast results in a high number of misses, which directly explains the significantly higher recall scores for Global MetNet. Simultaneously, Global MetNet’s ability to correctly forecast the storm’s location and intensity without generating widespread spurious precipitation accounts for its large gains in precision and overall skill as measured by CSI. The case study serves as a compelling example of an event where HRES has virtually no skill, while Global MetNet provides a highly accurate and useful forecast.

In conclusion, both the statistical and case-study analyses demonstrate that Global MetNet represents a significant advancement over HRES for short-term quantitative precipitation forecasting, particularly for challenging, high-impact weather events like convective systems.

\subsection{Mesoscale Convective Systems over ITCZ (East Africa)}

\begin{figure}[h!]
    \centering
    % Subfigure 1: Set width to less than half the text width
    \begin{subfigure}[b]{0.44\textwidth}
        \centering
        % Scale the image to the width of its subfigure container
        \includegraphics[width=\textwidth]{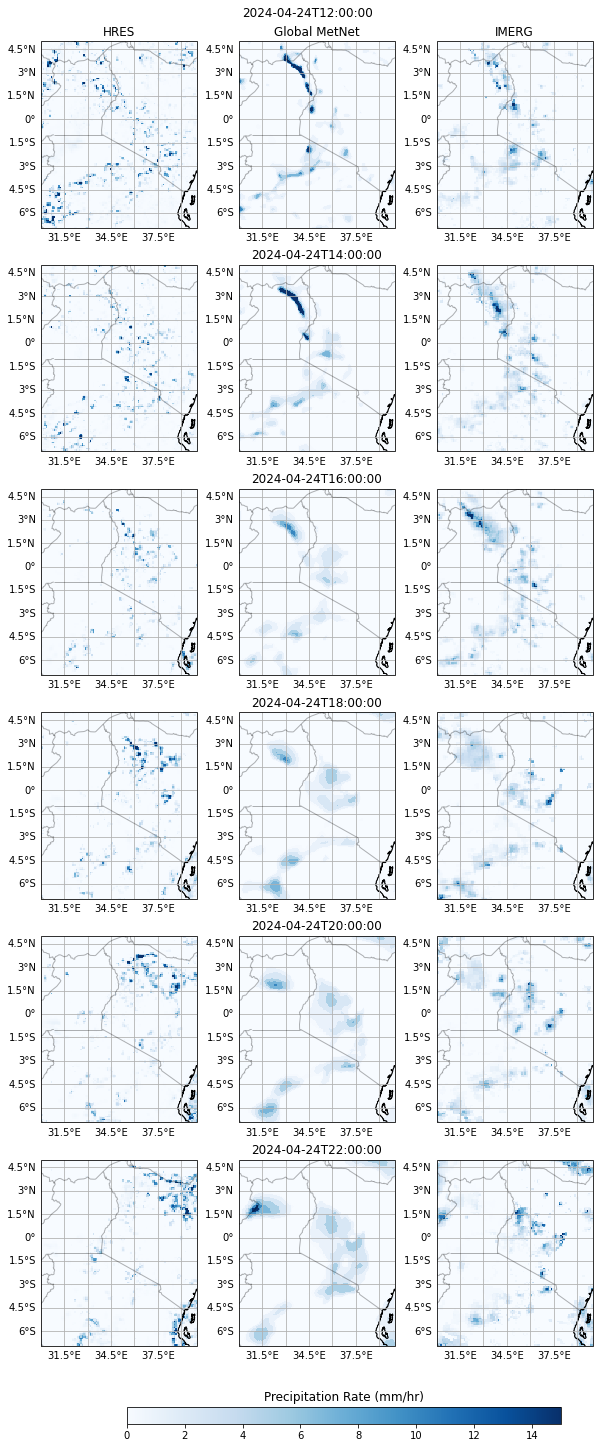}
        \caption{Forecast visualized for HRES (left), Global MetNet (middle) and IMERG (right).}
        \label{fig:DCS}
    \end{subfigure}
    \hspace{1em} % Adds horizontal space between the figures
    % Subfigure 2: Set width to less than half the text width
    \begin{subfigure}[b]{0.44\textwidth}
        \centering
        \includegraphics[width=\textwidth]{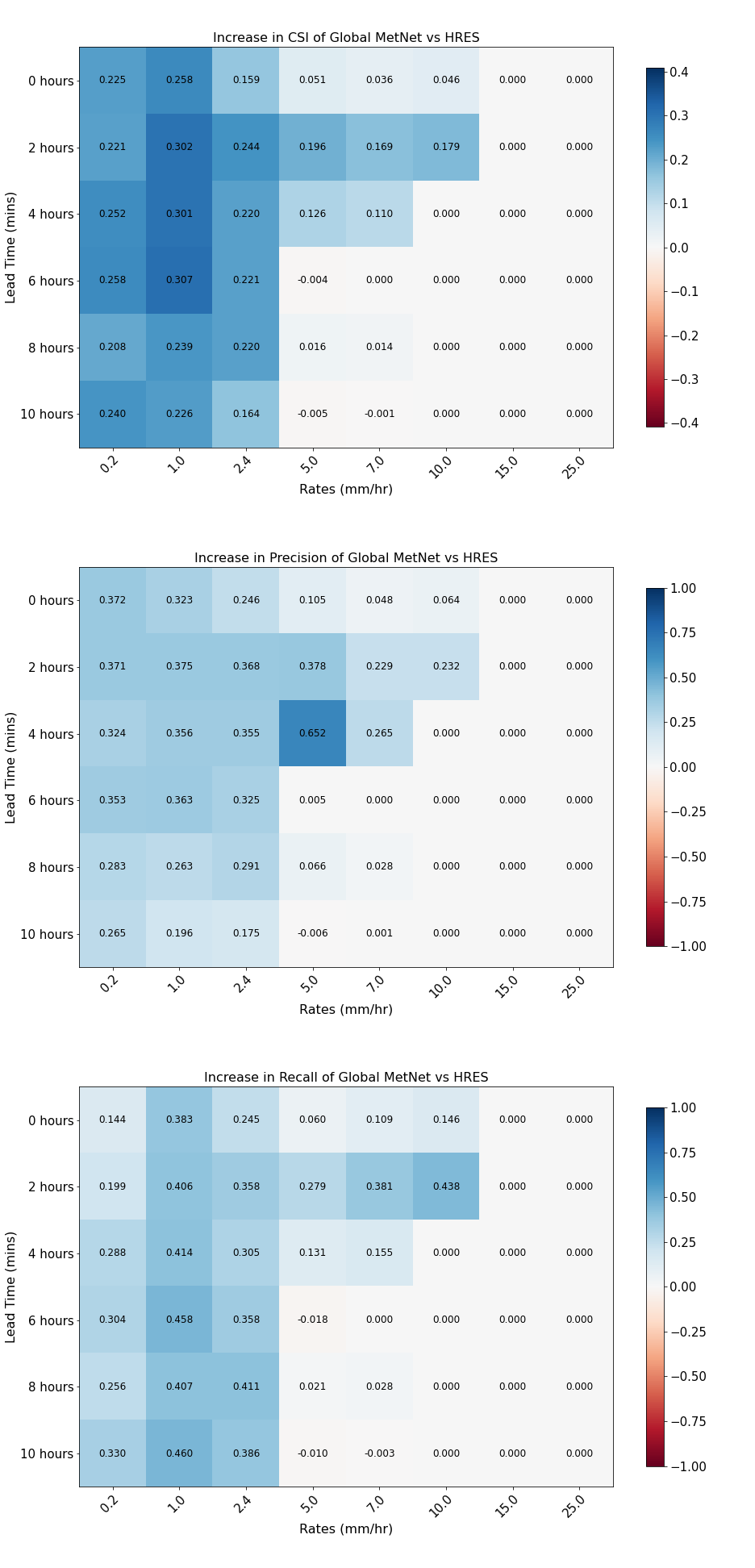}
        \caption{The difference in CSI, Precision and Recall  of the Global MetNet forecast over HRES forecast per lead time and per rate. Positive values imply an increase in metric and negative values imply a decrease.}
        \label{fig:DCS_stats}
    \end{subfigure}
    % A main caption describing the whole figure
    \caption{Case study of a diurnal convective system over East Africa showing nowcasts available at 12:00 UTC, April 24, 2024 leading up to 22:00 UTC.}
    \label{fig:dcs_analysis}
\end{figure}

On 24 April 2024, a north/south oriented mesoscale convective system (MCS) developed in eastern Uganda, shown in figure \ref{fig:dcs_analysis}. Within the MCS there were multiple regions of moderate to strong convection from 12-18 UTC. Over the day, this MCS moved west, weakening in evening hours from the loss of diurnal heating. Convection along the ITCZ is difficult for weather models because it is weakly forced and fairly transient in nature, which can be seen in the HRES output as widespread scattered precipitation with less coherence between consecutive two-hourly forecasts. This makes ITCZ convection an ideal location that benefits from nowcasting that incorporates observational datasets. The statistical analysis shows a similar increase in precision and CSI for Global MetNet as with the deep convective systems due to the improved prediction of location and intensity of the precipitation.

\subsection{Tropical Cyclones}

\begin{figure}[h!]
    \centering
    % Subfigure 1: Set width to less than half the text width
    \begin{subfigure}[b]{0.54\textwidth}
        \centering
        % Scale the image to the width of its subfigure container
        \includegraphics[width=\textwidth]{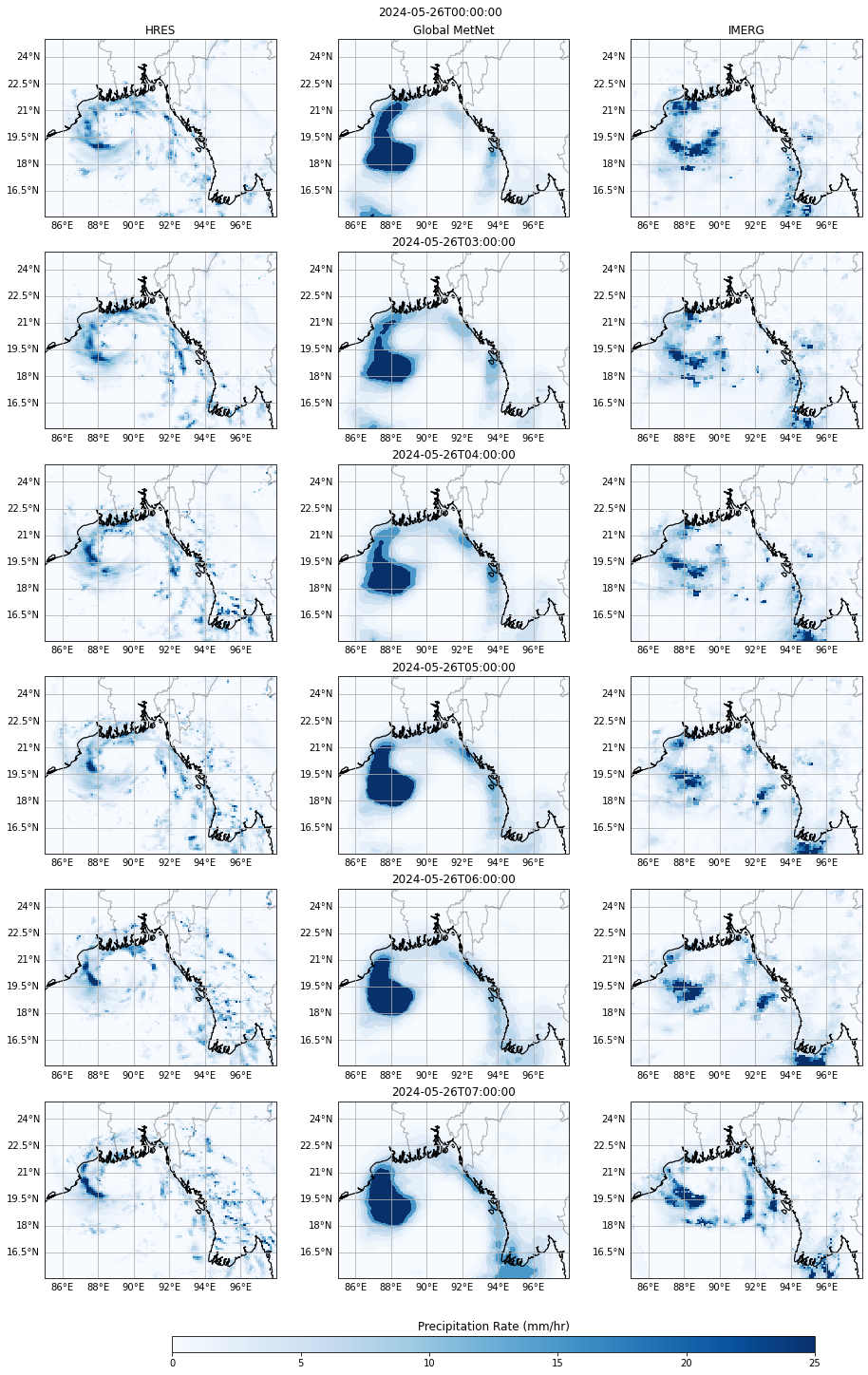}
        \caption{Cyclone Remal Nowcasts for HRES (left), Global MetNet (middle) and IMERG (right).}
        \label{fig:Remal}
    \end{subfigure}
    \hspace{1em} % Adds horizontal space between the figures
    % Subfigure 2: Set width to less than half the text width
    \begin{subfigure}[b]{0.34\textwidth}
        \centering
        \includegraphics[width=\textwidth]{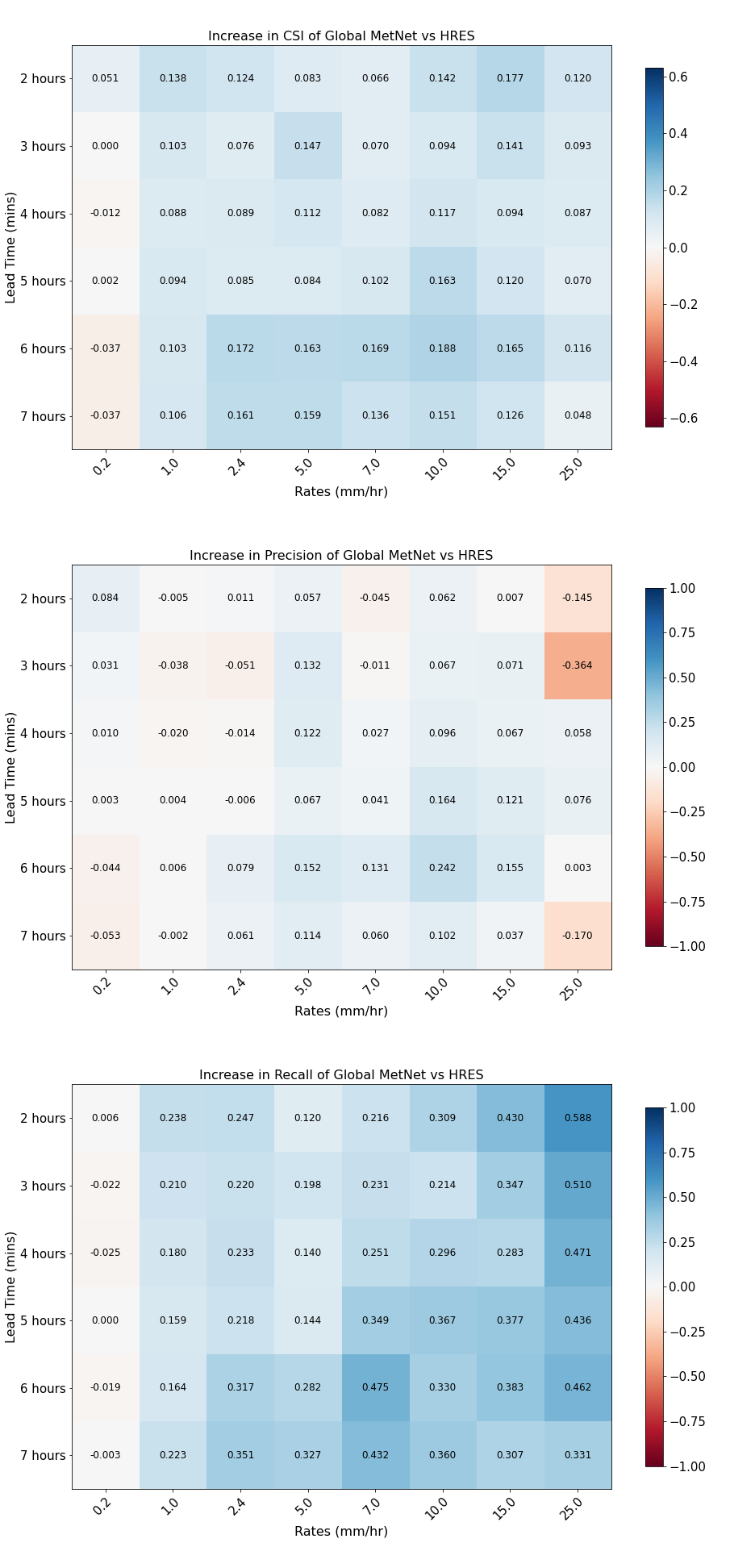}
        \caption{The difference in CSI, Precision and Recall  of the Global MetNet forecast over HRES forecast per lead time and per rate. Positive values imply an increase in metric and negative implies a decrease.}
        \label{fig:Remal_stats}
    \end{subfigure}
    % A main caption describing the whole figure
    \caption{Case study of Tropical Cyclone Remal over the Bay of Bengal showing nowcasts available at 00:00 UTC, May 26, 2024 leading up to 07:00 UTC.}
    \label{fig:remal_analysis}
\end{figure}

The following analysis assesses the performance of the Global MetNet and HRES models for a high-impact weather event, the Tropical Cyclone Remal in the Bay of Bengal. The evaluation reveals a critical trade-off between the models' forecast strategies, where Global MetNet's aggressive prediction of heavy rainfall results in superior overall skill despite a reduction in precision. The IMERG data shows a well-defined tropical cyclone with a distinct circulation and strong, curved rain bands containing embedded cores of very intense precipitation ($\geq$20 mm/hr). The HRES model captures the general location of the cyclone but severely underestimates the intensity of the rainfall, showing a diffused, smeared-out precipitation field with almost none of the high-intensity cores visible in IMERG. This explains its lower recall scores. Conversely, the overly broad precipitation shield from Global MetNet explains its lower precision scores. It correctly captures the heavy rain where it exists (high recall) but also incorrectly places heavy rain in the gaps between the actual rain bands (high false alarms). Note that, the HRES model was initialized at 18:00 UTC on May 25, 2024 whereas the Global MetNet model was initialized just a few mins prior to 00:00 UTC on May 26, 2024.

From a practical, hazard-forecasting perspective, MetNet's performance is arguably more valuable. Its high recall ensures that the life-threatening risk of extreme rainfall is not missed. The HRES forecast, while generating fewer false alarms, fails to convey the true severity of the event.

\section{Conclusion}
\label{sec:conclusion}
The work presented here introduces Global MetNet, an operational, deep-learning-based system for high-resolution precipitation nowcasting that demonstrates a significant leap forward in global forecast equity. By primarily leveraging geostationary satellite imagery and the GPM CORRA dataset, Global MetNet overcomes the critical limitations of traditional models, which are heavily reliant on ground-based radar infrastructure that is sparse in the Global South.

Our results show that Global MetNet consistently outperforms industry-standard NWP models, like HRES and HRRR, across all tested lead times and precipitation intensities. Crucially, it dramatically improves forecast skill in the tropics and other data-sparse regions, effectively closing the long-standing accuracy gap between the Global North and Global South. The model produces forecasts at an approximate 0.05° spatial and 15-minute temporal resolution for the next 12 hours, with an operational latency of under a minute, making it a powerful tool for real-world applications.

Despite these advances, we acknowledge certain limitations. The model's training in data-sparse regions relies on GPM CORRA as a proxy for ground truth, which itself is limited by satellite revisit times and therefore limits the amount of extreme precipitation data available for training. Furthermore, the model's tendency to over-predict intense rainfall—a `wet bias'—is preferable to under-prediction from a safety standpoint; however, it lacks the realistic looking spatial structures of such precipitation events. This highlights an area for further refinement, to provide sharper looking predictions in accurate locations without compromising on the intensity of the event.

This research represents a pivotal step towards democratizing access to accurate, life-saving weather information. Future work will focus on mitigating current limitations by exploring techniques to refine probabilistic forecasts and reduce biases in extreme events. We also aim to enhance the model by incorporating additional observational data sources, such as lightning activity, and to investigate pathways to make this technology and its outputs more accessible to meteorological agencies in developing nations. Through its current deployment to millions of users on Google Search, Global MetNet already demonstrates its operational readiness and practical value, paving the way for a future where AI-driven weather prediction can better serve communities worldwide.

\subsection{Acknowledgments}
We would like to thank Stephan Hoyer, Daniel Rothenberg, Rahul Mahrsee, Sahil Dabhi, Darshan Prajapati, Jash Rana, Piyush Ingale, Kunal Shah, Vusumzi Dube, Lasse Espeholt, Marcin Andrychowicz, Thomas Turnbull, Natalie Williams, Akib Uddin, Luke Barrington, Maya Ekron, Ariel Zevin, Maryam Tohidi, Isalo Montacute, Amanie Brik, Ashley Rivera, Carla Bromberg, and Jared Sisk for contributing in various capacities and roles towards this effort.

\bibliography{main}

\appendix
\section{Supplement}
\label{sec:supplement}

\subsection{Datasets}
\label{supp:datasets}
\subsubsection{GPM CORRA}
\label{supp:gpm2b}
To facilitate validation and training of the model on precipitation measurements from other parts of the world and especially the tropics, we make use of NASA’s Global Precipitation Measurement mission’s dual-frequency precipitation radar satellite.  This satellite also carries a microwave imager; we use the combined radar/radiometer retrieval leveraging both instruments as our primary evaluation target when terrestrial weather radar networks are not available. 

The Combined Radar-Radiometer Algorithm (CORRA) leverages both the dual-band Ka/Ku radars as well as the passive microwave imager (GMI) both carried by the GPM Core Observatory.  CORRA has a spatial footprint of roughly 250 km by 5 km per scan, and produces 8000 scans per orbit or $\sim$90 minutes.  The resulting data product has a $0.05^{\circ} \times 0.05^{\circ}$ spatial resolution;  a 15-minute segment from a given orbit covers a roughly 6500 km (along-track) $\times$ 250 km (cross-track) swath, with pixel dimensions of roughly 1300 (along-track) $\times$ 50 (cross-track). 

This data provides us with near-global coverage although the revisit rate of the satellite over a given area is only once every $\sim$2.5 days and a lower spatial resolution than ground weather radar data, which can make it difficult to resolve small-scale features.  Despite the limitations, GPM CORRA data is a valuable tool for tracking and forecasting precipitation. 

The following figure breaks down an example orbit into 15-minute segments, which is a reasonable time interval to treat as a “single image/observation”: 

\begin{figure}[h!]
    \centering
    \includegraphics[width=0.7\textwidth]{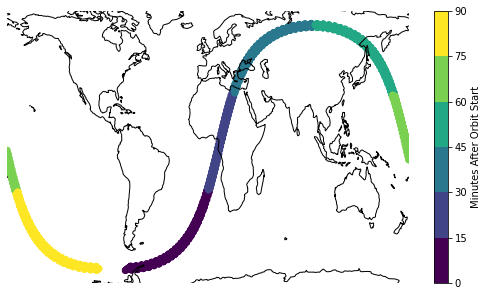}
    \caption{Sample GPM orbital swath color-coded by contiguous 15-minute intervals, using geolocation/timing data from the raw retrieval dataset. }
    \label{fig:supp_gpm_swath}
\end{figure}

\subsubsection{Radar}
\label{supp:radar}
As an auxiliary target, we leverage three sources of ground radar data that are precipitation estimates themselves on top of the raw radar measurements: 
\begin{itemize}
    \item MRMS data is a high-resolution, near-real-time product that provides precipitation estimates over the contiguous United States, using the National Oceanic and Atmospheric Administration (NOAA) National Weather Service (NWS) Next-Generation Radar (NEXRAD) network. 
    \item Opera radar data is a composite of radar data from several European countries, providing coverage over parts of Europe. 
    \item JMA radar data is a high-quality radar product from the Japan Meteorological Agency, providing coverage over Japan and surrounding areas. 
\end{itemize}

\subsubsection{Geostationary Satellite}
\label{supp:geosat}
The model utilizes data from a global constellation of geostationary satellites to provide multispectral imagery with various bands at high spatial and temporal resolutions and with a low real-time latency.  As described in the main paper, a total of 7 satellites are used to create mosaics at varying central wavelengths.  Here is how we determine the bands from different satellites to match together: 

Once we decided on which satellites to use, we had to come up with a way to decide which input bands to use for each mosaic.  We came up with the following algorithm. 
\begin{enumerate}
    \item Sort all bands for all satellites by the sensitivity width of each band in ascending order. 
    \item For each band in the sorted list: 
    \begin{itemize}
        \item If the band does not overlap with any bands on our list of chosen output bands, add this band to the list of chosen output bands. 
    \end{itemize}
    \item Define a mosaic for each of the chosen output bands. 
    \item For each band of each satellite: 
    \begin{itemize}
        \item Add the band to the mosaic that meets the following criteria: 
        \begin{itemize}
            \item The band has overlapping sensitivity with the mosaic’s wavelength range. 
            \item The center of the band is closer to the center of this mosaic’s range than any other mosaic. 
        \end{itemize}
    \end{itemize}
\end{enumerate}
This results in 18 mosaics with names and bands described in Table \ref{tab:mosaics}. 

\begin{table}[h!]
    \centering
    \caption{Mosaic names and contributing satellite bands. }
    \label{tab:mosaics}
    \resizebox{\textwidth}{!}{%
    \begin{tabular}{llll}
        \toprule
        \textbf{Mosaic Name} & \textbf{Central Wavelength} & \textbf{Wavelength Range} & \textbf{Contributing Satellites} \\
        \midrule
        mosaic\_470\_nm & 0.47 \textmu m & 0.45-0.49 \textmu m & Himawari-9, GOES-19, GOES-18, GK-2A, Meteosat-12 \\
        mosaic\_509\_nm & 0.509 \textmu m & 0.495-0.523 \textmu m & Himawari-9, GK-2A, Meteosat-12 \\
        mosaic\_640\_nm & 0.64 \textmu m & 0.62-0.66 \textmu m & Meteosat-11, Meteosat-9, Himawari-9, GOES-19, GOES-18, GK-2A, Meteosat-12 \\
        mosaic\_863\_nm & 0.863 \textmu m & 0.846-0.88 \textmu m & Meteosat-11, Meteosat-9, Himawari-9, GOES-19, GOES-18, GK-2A, Meteosat-12 \\
        mosaic\_914\_nm & 0.914 \textmu m & 0.894-0.934 \textmu m & Meteosat-12 \\
        mosaic\_1370\_nm & 1.37 \textmu m & 1.363-1.377 \textmu m & GOES-19, GOES-18, GK-2A, Meteosat-12 \\
        mosaic\_1610\_nm & 1.61 \textmu m & 1.59-1.63 \textmu m & Meteosat-11, Meteosat-9, Himawari-9, GOES-19, GOES-18, GK-2A, Meteosat-12 \\
        mosaic\_2250\_nm & 2.25 \textmu m & 2.225-2.275 \textmu m & Himawari-9, GOES-19, GOES-18, Meteosat-12 \\
        mosaic\_3830\_nm & 3.83 \textmu m & 3.74-3.92 \textmu m & Meteosat-11, Meteosat-9, Himawari-9, GOES-19, GOES-18, GK-2A, Meteosat-12 \\
        mosaic\_6200\_nm & 6.2 \textmu m & 6.0-6.4 \textmu m & Meteosat-11, Meteosat-9, Himawari-9, GOES-19, GOES-18, GK-2A \\
        mosaic\_6900\_nm & 6.94 \textmu m & 6.74-7.14 \textmu m & Himawari-9, GOES-19, GOES-18, GK-2A \\
        mosaic\_7330\_nm & 7.33 \textmu m & 7.24-7.42 \textmu m & Meteosat-11, Meteosat-9, Himawari-9, GOES-19, GOES-18, GK-2A, Meteosat-12 \\
        mosaic\_8590\_nm & 8.59 \textmu m & 8.415-8.765 \textmu m & Meteosat-11, Meteosat-9, Himawari-9, GOES-19, GOES-18, GK-2A, Meteosat-12 \\
        mosaic\_9620\_nm & 9.61 \textmu m & 9.42-9.8 \textmu m & Meteosat-11, Meteosat-9, Himawari-9, GOES-19, GOES-18, GK-2A, Meteosat-12 \\
        mosaic\_10400\_nm & 10.4 \textmu m & 10.2-10.6 \textmu m & Himawari-9, GOES-19, GOES-18, GK-2A, Meteosat-12 \\
        mosaic\_11200\_nm & 11.2 \textmu m & 11.0-11.4 \textmu m & Meteosat-11, Meteosat-9, Himawari-9, GOES-19, GOES-18, GK-2A \\
        mosaic\_12400\_nm & 12.4 \textmu m & 12.2-12.6 \textmu m & Meteosat-11, Meteosat-9, Himawari-9, GOES-19, GOES-18, GK-2A, Meteosat-12 \\
        mosaic\_13300\_nm & 13.3 \textmu m & 13.1-13.5 \textmu m & Meteosat-11, Meteosat-9, Himawari-9, GOES-19, GOES-18, GK-2A, Meteosat-12 \\
        \bottomrule
    \end{tabular}
    }
\end{table}

\subsubsection{Numerical Weather Prediction Dataset}
\label{supp:nwp}
The ECMWF High-Resolution forecast (HRES), a global NWP model that provides a single deterministic forecast of the weather, is used as an additional input where we use both the analysis and the hourly forecasts.  This data is initialized every 6 hours, and therefore when we initialize our own model at arbitrary times within the 6 hours, we use the last available analysis and forecast.  This means that there are initialization times of our model using up to a 12 hour old dataset from HRES. 

For the analysis dataset we use several surface and atmospheric variables.  We also take the surface variables forecasted for the initialization time of our model as additional input to provide the model with a forecasted current state of the world as an input.  Additionally, we take the total precipitation and precipitation rate forecasts made at the last initialization time of HRES for all the lead times of our model as an input, thereby informing our model of what has been forecasted by another model. 

\subsubsection{IMERG Final}
\label{supp:imerg}
The IMERG Final dataset provides precipitation estimates on a global scale with a spatial resolution of 0.1° x 0.1° and a temporal resolution of half an hour.  Due to its accuracy and reliability, it is considered a good source of precipitation estimates, valuable for validating and calibrating other precipitation products.  However, it is important to note that IMERG Final has a latency of approximately 3.5 months, which may limit its use in real-time applications. 

\subsubsection{Topography}
\label{supp:topo}
In addition to satellite and NWP data, we included topographical features such as elevation, latitude, and longitude which can enhance the model's predictive capabilities.  These features provide crucial context about the geographical characteristics of an area, which can influence precipitation patterns.  Elevation data helps the model understand how altitude affects precipitation.  For instance, orographic lift—where air is forced upwards by terrain—can lead to increased precipitation on the windward side of mountains.  By incorporating elevation data, the model can better predict these localized precipitation enhancements. 

\subsection{Metrics and Baselines}
\label{supp:metrics}
We use the following key metrics to evaluate across all precipitation rates and lead times to provide a comprehensive assessment of the model's performance under different conditions. 
\begin{itemize}
    \item \textbf{Critical Success Index (CSI):} This metric is commonly used in weather forecasting to measure the accuracy of predicting events like precipitation.  It focuses on correctly predicted events (hits) relative to the total number of actual events and false alarms.  A high CSI indicates that the model is good at predicting events without generating too many false alarms. 
    \item \textbf{Frequency Bias:} This metric compares the frequency of predicted events to the frequency of observed events.  A bias of 1 indicates that the model predicts events with the same frequency as they occur.  A bias greater than 1 means the model overpredicts events, while a bias less than 1 indicates underprediction. 
    \item \textbf{Fractions Skill Score (FSS):} This metric assesses the spatial distribution and intensity of predicted events compared to observed events.  It considers the fraction of grid points where the predicted and observed values exceed a certain threshold.  A high FSS indicates that the model accurately captures the spatial distribution and intensity of events. 
\end{itemize}

We evaluate our model against baselines, all at $0.05^{\circ} \times 0.05^{\circ}$ and 15 min temporal resolution.  If the baseline is available only at a lower spatial resolution then we upsample the baseline to match the resolution of Global MetNet’s forecast.  HRES (ECMWF High-Resolution Ensemble System) serves as one of the baselines for precipitation nowcasting evaluation due to its global coverage and provision of both analysis and hourly forecasts.  It offers a standard against which to measure the performance of new models.  Since HRES has an hourly temporal resolution, we only run validation at the top of the hour and assume that the hourly precipitation rate predicted by HRES will uniformly occur through the hour and therefore assume the first 15 minutes to have that precipitation rate. 

In addition to using HRES as a baseline, we created a post-processed version of HRES by training an ML model to predict GPM CORRA targets.  The ML model has the same architecture as the Global MetNet model but with only the HRES atmospheric and surface variables as input.  This helps calibrate HRES against CORRA and makes for a much stronger baseline, making it clear that any improvements upon this are due to other data sources used in our model.  This hybrid approach leverages the strengths of both NWP and AI models providing a more challenging and relevant comparison point.  It demonstrates the incremental value of our model along with the remote-sensed datasets beyond what a well-regarded existing system can achieve. 

\subsection{Ablation Studies}
\label{supp:results}
% \subsubsection{Additional Evaluations}
% \begin{figure}[h!]
%     \centering
%     \includegraphics[width=\textwidth]{figure_supplement_freq_bias_usa.png}
%     \caption{Frequency Bias of Global MetNet vs. NWP Baselines in the US (vs. MRMS) for Precipitation Rates of 0.2, 2.4, 7.0, and 25.0 mm/hr.}
%     \label{fig:supp_freq_bias_usa}
% \end{figure}

\begin{figure}[h!]
    \centering
    \includegraphics[width=\textwidth]{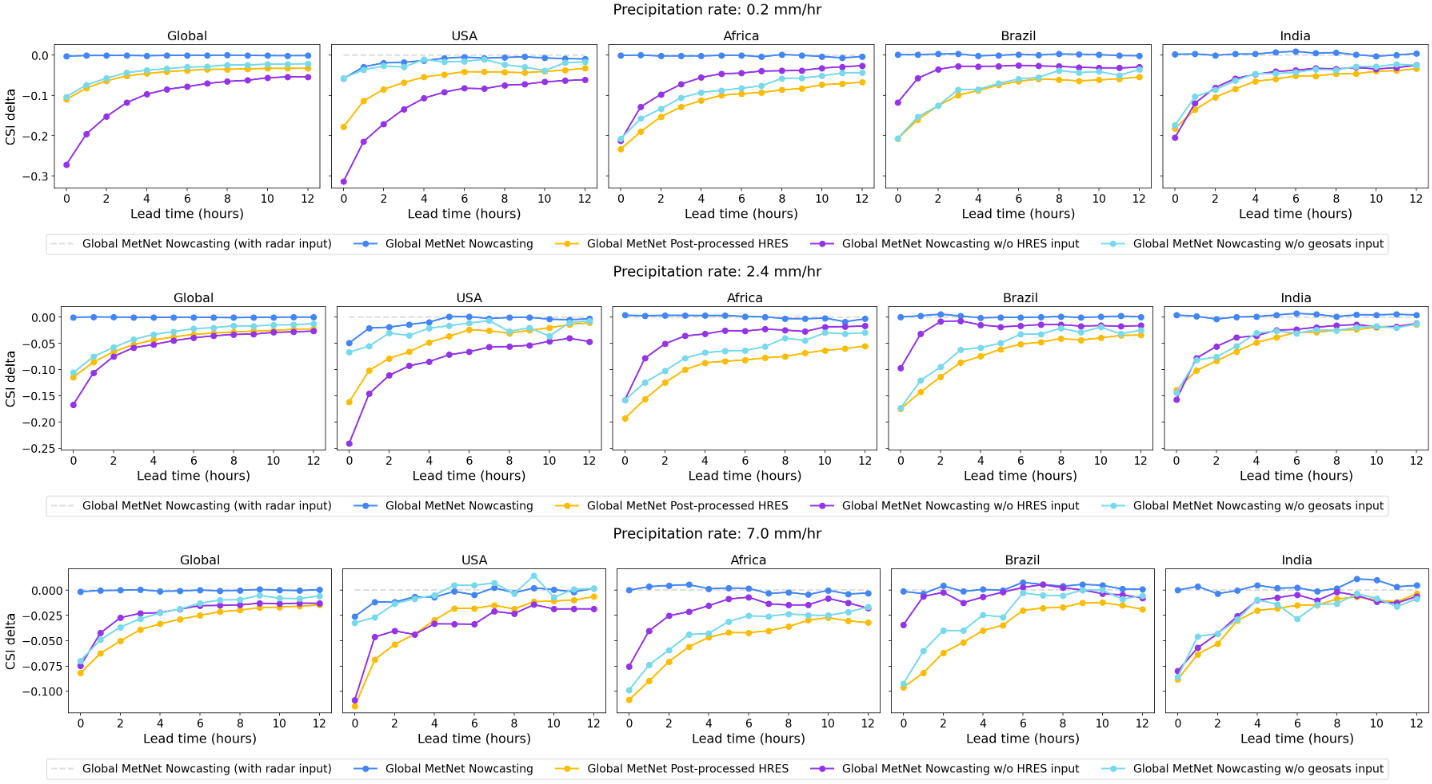}
    \caption{CSI delta from ablating various input datasets.}
    \label{fig:supp_ablation}
\end{figure}

To understand the contribution of each input and training target to the Global MetNet model's performance, we conducted ablation studies.  These studies involved systematically removing specific inputs or targets and observing the resulting changes in model metrics.  First, let’s look at the ablations on the input datasets in Figure \ref{fig:supp_ablation}. 

\subsubsection{Removing Geostationary Satellite Mosaics}
When we removed the geostationary satellite data (GOES-16, Himawari, Meteosat), there was a significant drop in the model's ability to predict fine-grained precipitation patterns, especially at shorter lead times.  The Critical Success Index (CSI) decreased the most among all other input ablations, by approximately 0.03 - 0.05 CSI points globally for estimating the current rate of precipitation.  This indicates that satellite data is crucial for capturing real-time atmospheric conditions and improving immediate forecast accuracy.This data also has the lowest real-time latency, another reason why it is such an important input into a nowcasting system.  Although as we get to the later lead times not having geostationary data did not seem to have too much of an impact on model performance. 

\subsubsection{Removing NWP Analysis and Forecasts}
Removing the Numerical Weather Prediction (NWP) data (HRES atmospheric and surface variables) had almost the opposite effect to removing the geostationary satellite datasets.  It led to a noticeable decline in performance, but particularly at longer lead times (beyond 6 hours) and had less of an effect at the immediate or shorter lead times.  The model's ability to predict large-scale precipitation systems and their evolution was reduced.  The CSI decreased by about 0.08 at 2.4 mm/hr precipitation rate globally, highlighting the importance of NWP data for contextualizing longer-term weather patterns. 

\subsubsection{Removing Ground Radar}
The contribution of radar data as an input feature can be evaluated by comparing the standard Global MetNet model against the variant trained with radar inputs. The results consistently demonstrate that including ground radar significantly boosts forecast accuracy, but this effect is geographically confined to regions where radar data is provided (USA, Europe, and Japan). The performance gain is most pronounced at shorter lead times (under 8 hours), highlighting the value of real-time radar observations for immediate nowcasting precision. In radar-sparse regions, the model's performance is identical with or without this input, confirming its robust baseline capability using satellite and NWP data for global coverage.

\subsubsection{Removing IMERG Early}
The model was also trained without the IMERG Early dataset as an input feature. This change produced no discernible impact on the model's predictive performance across any region, precipitation rate, or lead time. The lack of effect is attributed to the dataset's high real-time latency of 5 to 6 hours. By the time this data becomes available for inference, its informational value is superseded by the more timely data from lower-latency sources like the geostationary satellite mosaics, making it redundant as an input for the immediate nowcasting task.
% \paragraph{Removing GPM CORRA Data}
% Ablating the GPM CORRA data as a training target resulted in a substantial decrease in model performance in data-sparse regions.  The CSI in tropical regions dropped by approximately 0.15 at 0.2 mm/hr precipitation rate.  This underscores the significance of GPM CORRA data for providing ground truth in areas where radar data is unavailable.  Without this data, the model struggled to learn accurate precipitation patterns in these regions. 

% \paragraph{Summary of Ablation Study Results}
% \begin{table}[h!]
%     \centering
%     \caption{Summary of ablation study results. }
%     \label{tab:ablation_summary}
%     \begin{tabular}{p{3cm} p{6cm} p{2cm} p{2cm}}
%         \toprule
%         \textbf{Removed Dataset} & \textbf{Impact on Model Performance} & \textbf{Precipitation Rate (mm/hr)} & \textbf{Region} \\
%         \midrule
%         Geostationary Satellite Data & Significant drop in fine-grained prediction, especially at shorter lead times. & 0.2 & Globally \\
%         NWP Data & Reduced ability to predict large-scale systems, particularly at longer lead times. & 2.4 & Globally \\
%         GPM CORRA Data & Substantial decrease in performance in data-sparse regions, especially in the tropics. & 0.2 & Tropical \\
%         \bottomrule
%     \end{tabular}
% \end{table}

These ablation studies confirm that each input and training target plays a vital role in the Global MetNet model's overall performance.  The satellite data provides real-time observations, the NWP data offers contextual information, the GPM CORRA data serves as crucial ground truth in data-sparse regions, and the topographical data enhances the model's understanding of geographical influences on precipitation. 

\subsection{Supporting Figures}
\label{supp:sup_figs}

\begin{figure}[htbp]
    \centering
    \includegraphics[width=0.8\textwidth]{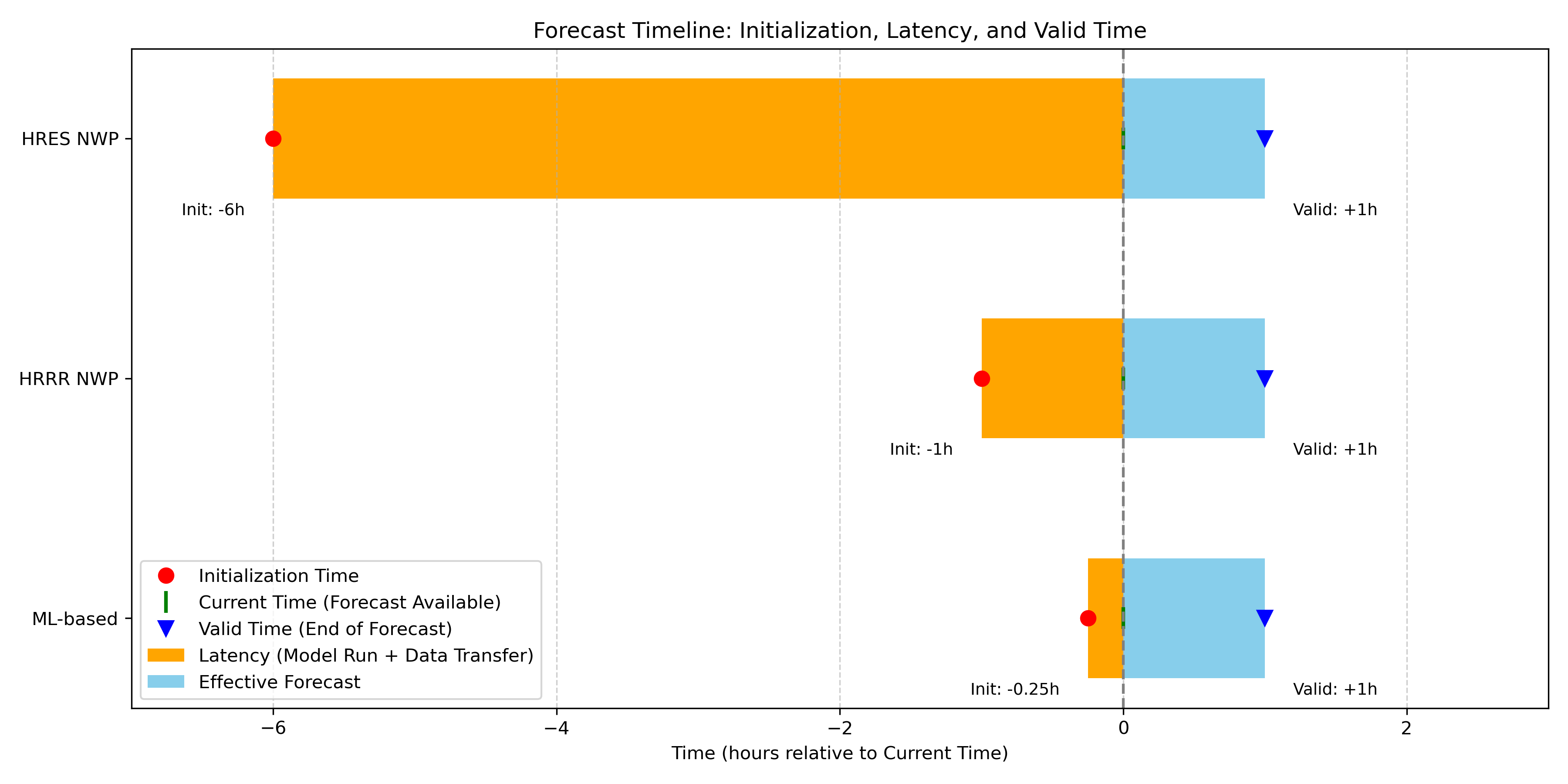}
    \caption{This timeline illustrates the process to get a forecast that is valid for the next hour (from `Current Time' to `+1h').  \textbf{Initialization Time (Red Circle):} This is when the model run begins. This shows how much earlier the HRES model must be initiated (6 hours before you need the forecast) compared to the ML model (15 minutes before).  \textbf{Latency (Orange Bar):} This is the time it takes for the model to run and for the data to become available.  It is the gap between the initialization and when the forecast becomes available.  \textbf{Current Time (Green Line):} This represents the moment the forecast is available.  \textbf{Effective Forecast (Blue Bar):} This is the 1-hour period for which the forecast is valid.  \textbf{Valid Time (Blue Triangle):} This marks the end of the 1-hour forecast period.  This visualization makes it clear how a long latency requires a much earlier start for the forecast to be useful for a specific time, highlighting the operational advantages of lower-latency models like the ML-based forecast. }
    \label{fig:timeline}
\end{figure}

\end{document}